\documentclass[letterpaper]{article} 
\usepackage{aaai2027}  
\usepackage[hyphens]{url}  
\usepackage{graphicx} 
\usepackage{natbib}  
\usepackage{caption} 
\usepackage{algorithm}
\usepackage{algorithmic}
\usepackage{amsmath}
\usepackage{amssymb}
\usepackage{booktabs}
\usepackage{multirow}
\usepackage[table]{xcolor}
\definecolor{TableBlue}{RGB}{46,117,145}
\definecolor{TableGold}{RGB}{184,80,40}
\definecolor{HeaderBlue}{RGB}{232,242,247}
\definecolor{AverageGold}{RGB}{250,239,231}
\definecolor{MethodGray}{gray}{0.92}
\definecolor{PendingGray}{gray}{0.96}
\newcommand{\deltarowshade}[1]{%
  \rlap{\smash{\begingroup\color{#1}\rule[-0.85ex]{\textwidth}{3.0ex}\endgroup}}%
}
\newcommand{\grouprowshade}[1]{%
  \rlap{\smash{\begingroup\color{#1}\rule[-0.3ex]{\textwidth}{2.74ex}\endgroup}}%
}

\usepackage{newfloat}
\usepackage{listings}
\DeclareCaptionStyle{ruled}{labelfont=normalfont,labelsep=colon,strut=off} 
\floatstyle{ruled}
\newfloat{listing}{tb}{lst}{}
\floatname{listing}{Listing}
\nocopyright

\title{Hi-TTRL: Regulating Consensus with Hints for Test-Time Reinforcement Learning}
\author{
    Kunbin Xu,
    Xingzuo Li,
    Xuefeng Bai,
    Kehai Chen
}
\affiliations{
    School of Computer Science and Technology, Harbin Institute of Technology, Shenzhen, China\\

    \{26B951004, 24S051028\}@stu.hit.edu.cn,
    \{chenkehai, baixuefeng\}@hit.edu.cn

}

\begin{document}

\maketitle

\begin{abstract}
Test-time reinforcement learning (TTRL) improves the reasoning capabilities of large language models without labeled data by updating the policy with pseudo-labels constructed through majority voting. While effective, the reward signal assigned from majority voting is highly sensitive to consensus strength, defined as the frequency of the most common answer within a rollout group. In TTRL, consensus strength plays a dual role: it reflects both the reliability of the pseudo-label and the distribution of advantages. Low consensus can amplify updates from unreliable pseudo-labels through disproportionately large advantages, whereas high consensus reduces reward contrast and ultimately yields vanishing gradients. In this paper, we introduce Hi-TTRL, a test-time reinforcement learning framework that utilizes hints during sampling to regulate rollout consensus strength. Hi-TTRL first estimates consensus strength from a partial rollout group. When the consensus strength falls outside a target interval, it invokes a Markov chain Monte Carlo (MCMC) hint sampler. The sampler targets the power-transformed prefix distribution and uses finite-step approximate sampling to generate rollout prefixes as hints. By tuning the power exponent, Hi-TTRL generates hints with a sharpened or flattened power target, steering rollout consensus strength toward the target interval. Experiments on multiple datasets and backbones show that Hi-TTRL consistently improves over standard TTRL, with ablations and consensus-steering analyses validating the effectiveness of adaptive hint-guided consensus regulation.
\end{abstract}


\section{Introduction}

Reinforcement learning (RL) has emerged as a highly effective paradigm for enhancing the complex reasoning capabilities of large language models (LLMs)\cite{guo2025deepseek, yang2025qwen3}. In particular, reinforcement learning with verifiable rewards (RLVR) provides a scalable way to optimize models on complex reasoning tasks\cite{shao2024deepseekmath}. RLVR leverages ground-truth annotations to provide models with deterministic and verifiable reward signals. Driven by these unambiguous reward signals, RLVR has yielded substantial improvements such as mathematical reasoning and code generation\cite{wang2026survey, yang2025code}.  Despite its effectiveness, RLVR still relies on a large amount of labeled and verifiable data. The acquisition of such high-quality, domain-specific annotations usually incurs prohibitive human and computational costs, making the training process difficult to scale to broader scenarios.

To alleviate the label dependency of RLVR, test-time reinforcement learning (TTRL) has recently emerged as a promising alternative\cite{zuo2025ttrl}. Instead of relying on external ground truth, TTRL dynamically constructs pseudo-labels by majority voting and assigns a rule based reward. Relying exclusively on these self-generated reward signals, the model's policy is then updated through the Group Relative Policy Optimization (GRPO) objective\cite{shao2024deepseekmath}. By internalizing the reward generation process, TTRL breaks the reliance on costly human annotations. 

\begin{figure}[t]
    \centering
    \includegraphics[width=1.0\columnwidth]{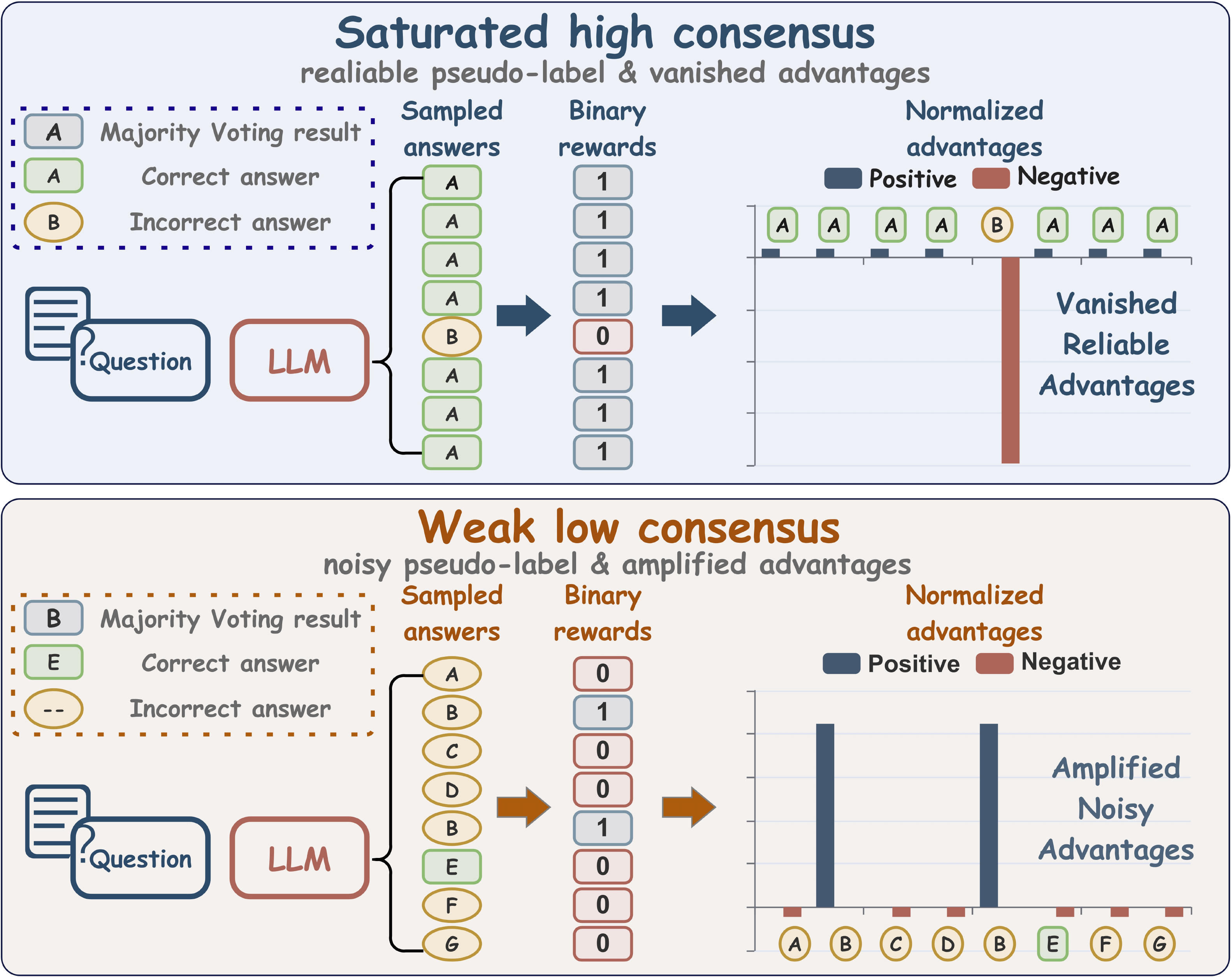} 
    \caption{Motivation for consensus-strength regulation in TTRL. Majority-voting rewards fail at both extremes: saturated consensus weakens gradients, while weak consensus can amplify spurious majorities.}
    \label{fig:motivation}
\end{figure}

Despite its promise, the efficacy of TTRL is constrained by an intrinsic coupling between the reliability of its pseudo-labels and the advantage estimation within the GRPO objective. We characterize this coupling through consensus strength, which measures  the frequency of the most common answer within a rollout group; we denote this quantity as $c(\mathcal{Y})$ and formalize it in Eq.~\eqref{eq:consensus_strength}. As illustrated in Figure~\ref{fig:motivation}, this coupling introduces a critical optimization paradox: under saturated high consensus, pseudo-labels are highly reliable, but uniform outcomes cause normalized advantages to vanish, depriving the policy of meaningful gradient signals. Conversely, under severely low consensus, majority-voted pseudo-labels are inherently noisy, yet the rare rollouts that match these spurious labels receive disproportionately large advantages, aggressively driving the policy toward flawed reasoning paths. Existing efforts to address this instability have largely explored the reward assignment and advantage estimation stages, employing techniques such as filtering pseudo-labels, reshaping rewards, or clipping advantages \cite{yan2026consensus, liu2025ettrl}. While these mechanisms are effective, regulating the consensus strength directly during the sampling stage remains unexplored.

To address this, we propose Hi-TTRL, a novel test-time reinforcement learning framework that explicitly regulates consensus strength during the sampling stage via an adaptive hint-generation mechanism. Instead of passively accepting the initial rollout distribution, Hi-TTRL actively monitors consensus strength online by evaluating an initial half-batch of sampled reasoning paths. When the resulting consensus strength falls outside a stable target interval, Hi-TTRL generates prefix hints with a Markov chain Monte Carlo (MCMC) sampler constructed from the current policy whose finite-step transitions approximately target a power-transformed distribution~\cite{karan2025reasoning}. The shape of the distribution is tightly governed by a dynamic exponent $\alpha$: under low consensus strength, an exponent $\alpha>1$ sharpens the target to encourage output convergence; conversely, under high consensus strength, an exponent $\alpha<1$ flattens the distribution to promote exploratory divergence. The current policy model is then utilized to complete these MCMC-sampled hints. By proactively intervening before the final voting and reward assignment, Hi-TTRL effectively steers the rollout distribution toward an optimal consensus-strength range, mitigating both noisy pseudo-labels and vanishing advantages.

Our main contributions are summarized as follows:
\begin{itemize}
    \item We identify consensus strength extremes as a source of instability in TTRL, which may contribute to amplifying noisy pseudo-labels and vanishing policy gradients.


    \item We adopt an adaptive MCMC hint sampler that relies solely on the old policy while approximately targeting a power-transformed prefix distribution.

    \item We propose Hi-TTRL, a test-time reinforcement learning framework that proactively regulates the consensus strength before majority voting and reward assignment.
\end{itemize}

\section{Related Works}

\noindent\textbf{Reinforcement Learning for Reasoning.}
Reinforcement learning has become a central paradigm for improving the reasoning and instruction-following abilities of LLMs. RLHF aligns models with human preferences through annotated comparisons and learned reward models \cite{ouyang2022training,jiang2024survey}, and has also been connected theoretically to contrastive learning \cite{lv2025hiddenlinkrlhfcontrastive}, with policy and preference optimization methods such as PPO, DPO, and SimPO \cite{schulman2017ppo,rafailov2023dpo,meng2024simpo}. More recently, RLVR replaces preference rewards with deterministic outcome signals and has proven effective for mathematical and code reasoning \cite{shao2024deepseekmath,openai2024o1,guo2025deepseek,yang2025qwen3,wang2026survey,yang2025code}. It has also motivated group-relative objectives such as GRPO \cite{shao2024deepseekmath}, reward design with synthetic or partially reliable data and meta-rewards \cite{gao2024reward,setlur2024incorrect,liang2026dual}, and methods that improve sample efficiency or credit assignment through multi-token reasoning units, informative one-shot examples, self-generated hints, high-entropy minority tokens, process--outcome reward harmonization, or adversarial critics \cite{xu2026beyondtoken,wang2025oneexample,qiyuan2026hipo,wang2025highentropy,ye2025beyondcorrectness,wu2025rlac}. However, most RLVR-style methods still rely on ground-truth answers, external verifiers or curated examples, motivating test-time reinforcement learning where rewards must be constructed from the model's own rollouts.

\noindent\textbf{Unsupervised Reinforcement Learning and TTRL.}
Test-time reinforcement learning (TTRL) extends unsupervised self-improvement to unlabeled test streams by sampling multiple rollouts, constructing pseudo-labels through majority voting, and updating the policy with pseudo-rewards \cite{wang2022selfconsistency,zuo2025ttrl}. Existing work improves pseudo-label quality either by augmenting majority voting with internal signals such as self-certainty or entropy-based rewards, reasoning topology, and distribution matching \cite{zhao2025learning,zhang2025empo,wang2026sarl,chen2026powerflow}, or by revising voting through self-play, self-rewriting, generator-verifier co-evolution, strict consensus criteria, negative pseudo-labeling, and confidence-weighted subgroup estimation \cite{wang2026selfharmony,yao2025selfrewriting,pan2026coverrl,yan2026consensus,wang2026scope}. Other methods mitigate early-stage noise by filtering low-support or uncertain samples, or by clipping and recalibrating advantages from noisy pseudo-labels \cite{zhang2025empo,huang2026rzero,liu2025ettrl,zhao2026echo}. These approaches mainly score, filter, revise, or reshape pseudo-labels after rollouts have been sampled. In contrast, Hi-TTRL targets the sampling stage, using power-transformed hints to steer consensus strength before final voting and reward assignment.

\begin{figure*}[t]
    \centering
    \includegraphics[width=\textwidth]{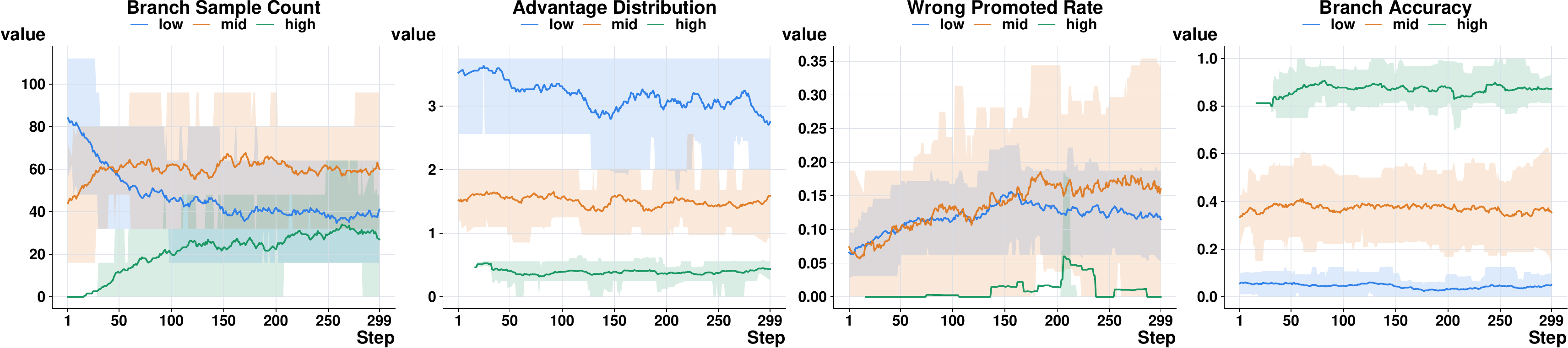}
    \caption{
    Diagnostic study of consensus-strength levels in majority-voting TTRL on AMC2023 with Qwen2.5-Math-1.5B.
    Prompt groups are partitioned into low-, mid-, and high-consensus branches.
    From left to right, the panels report the branch sample count, maximum positive normalized GRPO advantage, wrong-promoted rate, and ground-truth branch accuracy.
    Detailed metric definitions are provided in the Appendix.
    }
    \label{fig:preliminary_experiment}
\end{figure*}

\section{Preliminary Experiments}

Before introducing Hi-TTRL, we conduct a diagnostic pilot study to examine how different levels of consensus strength affect the pseudo-label reliability and gradient informativeness in real TTRL training. Specifically, we train Qwen2.5-Math-1.5B~\cite{yang2024qwen25math} on AMC2023~\cite{hfmathaiamc23}. At each training step, we sample a batch of 8 prompts, generate 32 rollouts per prompt for majority voting, and select 16 of these rollouts for the GRPO update. For each update rollout set, we first compute its consensus strength and then categorize it as low $[0,0.25)$, mid $[0.25,0.75)$, or high $[0.75,1]$. Figure~\ref{fig:preliminary_experiment} compares the training dynamics across the three consensus branches. The wrong-promoted rate measures the fraction of update rollouts within each branch that receive positive normalized advantages despite producing incorrect final answers. Ground-truth answers are used only for diagnostic evaluation, not reward construction.

As shown in Figure~\ref{fig:preliminary_experiment}, the low- and high-consensus branches exhibit opposite training dynamics. Low-consensus samples are frequent at the beginning of training and gradually decrease, but this branch consistently shows lower ground-truth accuracy, a higher wrong-promoted rate, and the largest maximum positive advantages. In contrast, high-consensus samples gradually become dominant, with high branch accuracy and near-zero wrong-promoted rate, yet their maximum positive advantages remain consistently small. These patterns reveal complementary limitations of the two extreme consensus levels. Low-consensus groups combine unreliable pseudo-labels with excessively large positive advantages, leading to strongly amplified incorrect updates, whereas high-consensus groups produce reliable pseudo-labels but insufficient reward contrast, leaving only weak gradient signals for GRPO optimization. This motivates Hi-TTRL to control consensus strength during sampling by promoting convergence for low-consensus groups and controlled divergence for high-consensus groups.

\begin{figure*}[t]
    \centering
    \includegraphics[width=0.9\textwidth]{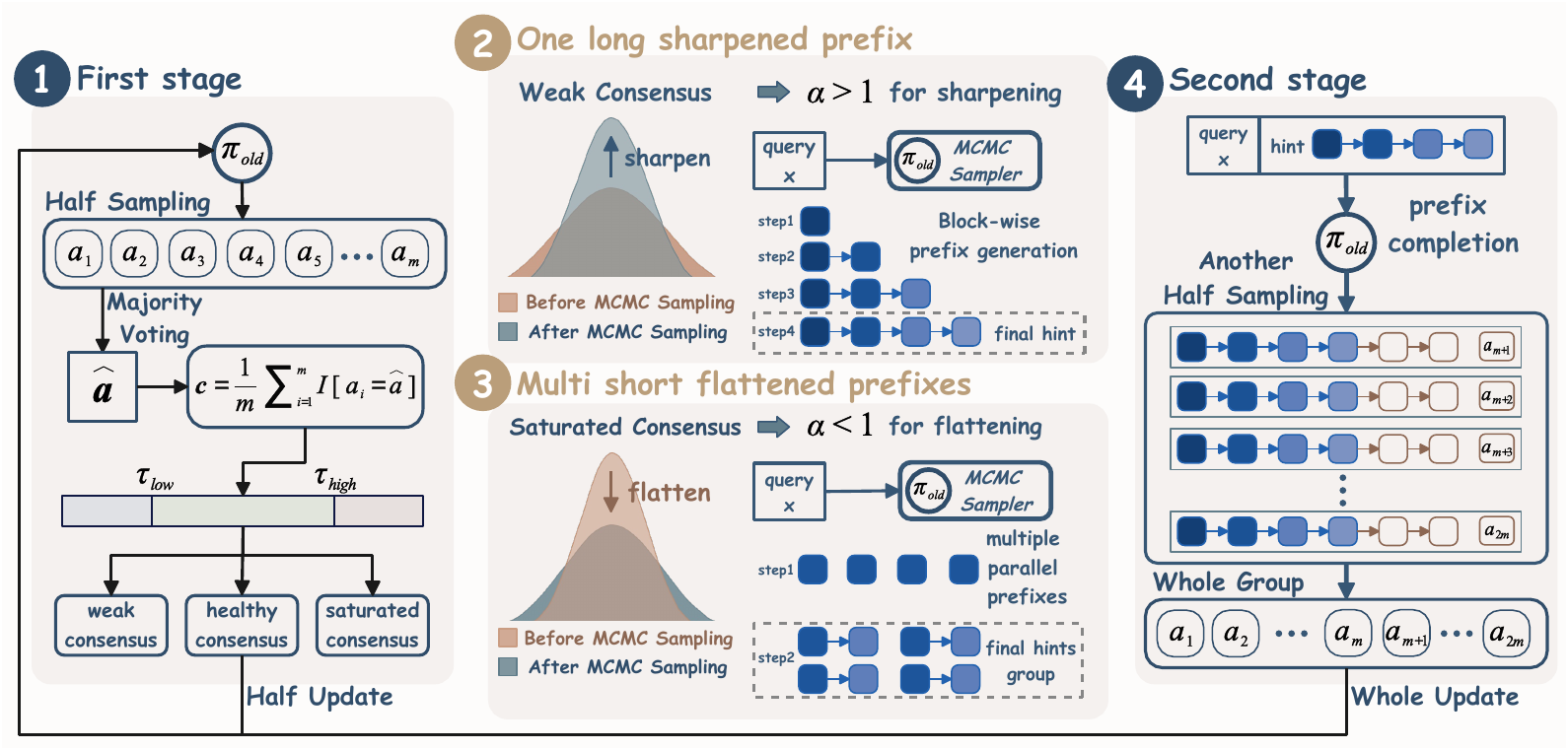}
    \caption{Overview of Hi-TTRL. Based on partial-rollout consensus, Hi-TTRL directly updates in-range groups and applies anchor or exploratory MCMC hints to low- or high-consensus groups.}
    \label{fig:method}
\end{figure*}

\section{Method}

\subsection{Preliminaries}

We adopt Group Relative Policy Optimization (GRPO) as the policy optimization backbone. Given a query $x$, the old policy $\pi_{\theta_{\mathrm{old}}}$ samples a rollout group $\mathcal{Y}=\{y_i\}_{i=1}^{m}$, where each rollout $y_i$ is mapped to a final answer $a_i=\operatorname{Ans}(y_i)$. The majority answer $\hat{a}$ is used as the pseudo-label, and its frequency defines the consensus strength:
\begin{equation}
\label{eq:consensus_strength}
\resizebox{0.88\linewidth}{!}{$\displaystyle
    c(\mathcal{Y})
    = \frac{1}{m}\sum_{i=1}^{m}\mathbb{I}[a_i=\hat{a}],
    \qquad
    \hat{a}
    = \arg\max_{a\in\mathcal{A}}\sum_{i=1}^{m}\mathbb{I}[a_i=a].
$}
\end{equation}
TTRL rewards rollouts that match $\hat{a}$ and normalizes their advantages within the group. Consequently, low consensus couples an unreliable pseudo-label with large advantages, while saturated consensus leaves little reward contrast. We retain the standard GRPO update~\cite{shao2024deepseekmath} and instead regulate the rollout distribution during sampling.

Following the solution-space view of StepHint~\cite{zhang2025stephint}, we regard autoregressive reasoning as a progressive exploration of candidate solutions: each generated prefix conditions subsequent decisions and prunes incompatible continuations. A prefix hint can therefore guide decoding toward a more promising subspace without specifying the complete reasoning trajectory. Its length controls the strength of this pruning: a longer prefix imposes more constraints and narrows the remaining search space more aggressively, whereas a shorter prefix provides lighter guidance and preserves greater freedom for exploration. This reduces the search difficulty while preserving flexibility for the model to complete the remaining reasoning.

To control the direction of this search, we consider power sampling, which draws a complete reasoning trajectory from $p_{\alpha}(y\mid x)\propto\pi_{\theta_{\mathrm{old}}}(y\mid x)^{\alpha}$. The power transformation preserves the ranking of trajectories but changes their relative probability gaps: $\alpha>1$ amplifies high-likelihood paths and concentrates search on a smaller set of directions, whereas $0<\alpha<1$ compresses these gaps and allocates more mass to alternative policy-supported paths. The exponent therefore provides direct control over the direction and breadth of search. \citet{karan2025reasoning} operationalize this target with an MCMC sampler that progressively resamples token subsequences. We adapt their construction from complete trajectories to prefixes, using power-targeted hints to steer the subspace explored by subsequent decoding.

\subsection{Hi-TTRL: Sampling-Stage Consensus Regulation}

To avoid the failure modes arising from extreme consensus strength, Hi-TTRL proactively regulates consensus strength at the sampling stage. As illustrated in Figure~\ref{fig:method}, Hi-TTRL first estimates consensus strength from a partial rollout group; groups within the target interval are updated directly, while low- or high-consensus groups trigger a power-target MCMC hint sampler that adds convergence- or divergence-oriented prefixes before the final vote and policy update.

Let $G$ denote the full rollout-group size when the second sampling stage is activated. For each query $x$, Hi-TTRL initially samples $G/2$ responses from the old policy to form the first-stage rollout group. It then extracts the answers, performs an intermediate majority vote, and computes the first-stage consensus strength using Eq.~\eqref{eq:consensus_strength}:
\begin{equation}
    c^{(1)}
    =
    c(\mathcal{Y}^{(1)}),
    \qquad
    \mathcal{Y}^{(1)}=\{y_i\}_{i=1}^{G/2},
\end{equation}
where $y_i$ denotes a response sampled from $\pi_{\theta_{\mathrm{old}}}(\cdot\mid x)$.

The consensus strength $c^{(1)}$ provides an online estimate of whether the current rollout group lies within a healthy consensus range. If $c^{(1)}$ falls inside a target interval $[\tau_{\mathrm{low}}, \tau_{\mathrm{high}}]$, Hi-TTRL regards the first-stage samples as exhibiting a reasonable trade-off between pseudo-label reliability and advantage diversity. In this case, no additional intervention is applied, and the policy is updated directly using the first-stage rollout group. When the first-stage consensus strength falls outside the target interval, Hi-TTRL triggers a second hint-guided sampling stage. The goal of this stage is to actively reshape the remaining rollout distribution before final voting and reward assignment. The second-stage rollouts are generated by the old policy conditioned on prefix hints produced by the power-target MCMC sampler. More details of the sampler will be described in the following subsection.

We finally define the rollout group used for the policy-gradient update as a consensus-adaptive composition of the two stages. Let the second-stage hint-guided completions produced by this sampler be
\begin{equation}
    \mathcal{Y}^{(2)}_{\mathrm{hint}}
    =
    \{y_i\}_{i=G/2+1}^{G}.
\end{equation}
The samples used to compute the final majority vote, rewards, GRPO advantages, and policy gradient are then
\begin{equation}
    \mathcal{Y}_{\mathrm{upd}}(x)
    =
    \begin{cases}
    \mathcal{Y}^{(1)},
    &
    \tau_{\mathrm{low}}
    \le
    c^{(1)}
    \le
    \tau_{\mathrm{high}},
    \\
    \mathcal{Y}^{(1)}
    \cup
    \mathcal{Y}^{(2)}_{\mathrm{hint}},
    &
    c^{(1)}
    <
    \tau_{\mathrm{low}}
    \ \text{or}\
    c^{(1)}
    >
    \tau_{\mathrm{high}}.
    \end{cases}
\end{equation}
Thus, the number of update rollouts is adaptive:
\begin{equation}
    n_{\mathrm{upd}}(x)
    =
    |\mathcal{Y}_{\mathrm{upd}}(x)|
    =
    \begin{cases}
    G/2, & c^{(1)}\in[\tau_{\mathrm{low}},\tau_{\mathrm{high}}],\\
    G, & c^{(1)}\notin[\tau_{\mathrm{low}},\tau_{\mathrm{high}}].
    \end{cases}
\end{equation}
The final majority vote and advantage normalization are always computed over $\mathcal{Y}_{\mathrm{upd}}(x)$. This adaptive update size is intentional: Hi-TTRL allocates additional samples only when the first-stage consensus falls outside the target interval.

\subsection{Power-Target MCMC Hint Sampling}

We adapt the power-sampling approximation of~\citet{karan2025reasoning} from complete responses to short prefixes. For a query $x$, the power-target hint distribution is
\begin{equation}
    \label{eq:power_hint_main}
    q_{\alpha}(h\mid x)
    \propto
    \pi_{\theta_{\mathrm{old}}}(h\mid x)^{\alpha},
\end{equation}
where $h$ denotes a partial reasoning prefix and the power exponent $\alpha$ controls the shape of the hint distribution. 

When $\alpha>1$, the distribution is sharpened: prefixes with higher likelihood under the rollout policy are further up-weighted, making second-stage rollouts more likely to share a coherent reasoning direction. This is useful under weak first-stage consensus, where stronger guidance is needed to promote rollout convergence. When $\alpha<1$, the distribution is flattened: lower-likelihood but still policy-supported prefixes receive relatively more mass, creating alternative reasoning branches. This is useful under saturated first-stage consensus, where the rollout group needs more diversity.

The hint is then concatenated with the original query and completed by the same rollout policy:
\begin{equation}
    y = [h;\tilde{y}],
    \qquad
    \tilde{y}\sim \pi_{\theta_{\mathrm{old}}}(\cdot \mid x,h).
\end{equation}
In this way, the MCMC approximation to the power target only determines the prefix-level steering signal, while the remaining response tokens are still generated by $\pi_{\theta_{\mathrm{old}}}$.

\begin{table*}[!t]
\centering
\begingroup
\small
\setlength{\tabcolsep}{2.9pt}
\renewcommand{\arraystretch}{1.12}
\newcommand{\greedyhead}{\shortstack{\texttt{greedy}}}
\newcommand{\meanhead}{\shortstack{\texttt{mean}}}
\begin{tabular*}{\textwidth}{@{\extracolsep{\fill}}l*{12}{c}@{}}
\toprule[1.2pt]
\multirow{2}{*}{\shortstack[l]{\textbf{Models}\\\textbf{/ Methods}}}
& \multicolumn{2}{c}{AMC-2023}
& \multicolumn{2}{c}{MATH-500}
& \multicolumn{2}{c}{AIME-2024}
& \multicolumn{2}{c}{MINERVA}
& \multicolumn{2}{c}{GAOKAO2023-en}
& \multicolumn{2}{c}{Average} \\
\cmidrule(lr){2-3}\cmidrule(lr){4-5}\cmidrule(lr){6-7}\cmidrule(lr){8-9}\cmidrule(lr){10-11}\cmidrule(lr){12-13}
& \greedyhead & \meanhead
& \greedyhead & \meanhead
& \greedyhead & \meanhead
& \greedyhead & \meanhead
& \greedyhead & \meanhead
& \greedyhead & \meanhead \\
\midrule[1.1pt]
\multicolumn{13}{@{}l@{}}{\grouprowshade{MethodGray}\makebox[\textwidth][c]{\textsc{I. Math-Specialized Backbone}}} \\
\multicolumn{13}{@{}l@{}}{\grouprowshade{PendingGray}\makebox[\textwidth][c]{\textsc{Qwen2.5-Math-1.5B}}} \\
Baseline & 33.73 & 27.26 & 32.04 & 34.32 & \underline{16.33} & 9.58 & 8.82 & 8.46 & 20.00 & 21.79 & 22.18 & 20.28 \\
TTRL & \underline{44.58} & \underline{44.88} & \underline{73.92} & \underline{71.35} & 13.67 & \underline{15.88} & 16.91 & 17.14 & 62.60 & \underline{62.11} & \underline{42.34} & \underline{42.27} \\
Intuitor & 37.35 & 36.45 & 70.80 & 67.63 & 11.67 & 9.29 & \underline{25.37} & \underline{22.04} & \underline{64.16} & 58.86 & 41.87 & 38.85 \\
\textbf{Hi-TTRL} & \textbf{51.81} & \textbf{48.12} & \textbf{78.00} & \textbf{76.86} & \textbf{24.33} & \textbf{18.25} & \textbf{36.76} & \textbf{35.96} & \textbf{70.13} & \textbf{69.17} & \textbf{52.21} & \textbf{49.67} \\
\deltarowshade{TableBlue!10}$\Delta$ & +7.23 & +3.24 & +4.08 & +5.51 & +10.66 & +2.37 & +19.85 & +18.82 & +7.53 & +7.06 & +9.87 & +7.40 \\
\deltarowshade{TableBlue!20}$\Delta\%$ & $\uparrow$16.22\% & $\uparrow$7.22\% & $\uparrow$5.52\% & $\uparrow$7.72\% & $\uparrow$77.98\% & $\uparrow$14.92\% & $\uparrow$117.39\% & $\uparrow$109.80\% & $\uparrow$12.03\% & $\uparrow$11.37\% & $\uparrow$23.31\% & $\uparrow$17.51\% \\
\midrule[1.1pt]
\multicolumn{13}{@{}l@{}}{\grouprowshade{MethodGray}\makebox[\textwidth][c]{\textsc{II. General Base Backbones}}} \\
\multicolumn{13}{@{}l@{}}{\grouprowshade{PendingGray}\makebox[\textwidth][c]{\textsc{Qwen3-1.7B-Base}}} \\
Baseline & 26.51 & 22.29 & 44.52 & 39.07 & 4.33 & 2.08 & 5.88 & 8.07 & 25.45 & 23.28 & 21.34 & 18.96 \\
TTRL & \underline{32.53} & \underline{36.37} & \underline{69.80} & \underline{69.17} & \underline{9.67} & \underline{7.50} & \underline{30.51} & \underline{30.63} & \underline{59.40} & \underline{59.38} & \underline{40.38} & \underline{40.61} \\
Intuitor & 30.12 & 26.51 & 56.80 & 56.65 & 4.33 & 4.58 & 24.26 & 22.52 & 49.87 & 48.17 & 33.08 & 31.69 \\
\textbf{Hi-TTRL} & \textbf{38.55} & \textbf{39.76} & \textbf{71.40} & \textbf{71.26} & \textbf{12.67} & \textbf{11.25} & \textbf{34.19} & \textbf{32.74} & \textbf{61.82} & \textbf{61.56} & \textbf{43.73} & \textbf{43.31} \\
\deltarowshade{TableBlue!10}$\Delta$ & +6.02 & +3.39 & +1.60 & +2.09 & +3.00 & +3.75 & +3.68 & +2.11 & +2.42 & +2.18 & +3.34 & +2.70 \\
\deltarowshade{TableBlue!20}$\Delta\%$ & $\uparrow$18.51\% & $\uparrow$9.32\% & $\uparrow$2.29\% & $\uparrow$3.02\% & $\uparrow$31.02\% & $\uparrow$50.00\% & $\uparrow$12.06\% & $\uparrow$6.89\% & $\uparrow$4.07\% & $\uparrow$3.67\% & $\uparrow$8.28\% & $\uparrow$6.66\% \\
\midrule
\multicolumn{13}{@{}l@{}}{\grouprowshade{PendingGray}\makebox[\textwidth][c]{\textsc{Qwen3-4B-Base}}} \\
Baseline & 24.10 & 18.75 & 34.72 & 31.89 & 13.33 & 7.50 & 12.13 & 13.40 & 14.55 & 12.73 & 19.77 & 16.85 \\
TTRL & 46.69 & 48.19 & \underline{81.68} & \underline{81.67} & \underline{21.67} & \textbf{21.04} & \underline{41.18} & \underline{42.37} & \textbf{71.17} & \textbf{71.10} & \underline{52.48} & \underline{52.87} \\
Intuitor & \underline{51.81} & \underline{53.09} & 74.40 & 74.92 & 9.33 & 9.52 & 29.04 & 29.50 & 69.09 & 68.39 & 46.73 & 47.08 \\
\textbf{Hi-TTRL} & \textbf{54.22} & \textbf{54.49} & \textbf{82.76} & \textbf{82.57} & \textbf{23.33} & \underline{20.83} & \textbf{44.12} & \textbf{43.27} & \underline{70.65} & \underline{70.65} & \textbf{55.02} & \textbf{54.36} \\
\deltarowshade{TableBlue!10}$\Delta$ & +7.53 & +6.30 & +1.08 & +0.90 & +1.66 & -0.21 & +2.94 & +0.90 & -0.52 & -0.45 & +2.54 & +1.49 \\
\deltarowshade{TableBlue!20}$\Delta\%$ & $\uparrow$16.13\% & $\uparrow$13.07\% & $\uparrow$1.32\% & $\uparrow$1.10\% & $\uparrow$7.66\% & $\downarrow$1.00\% & $\uparrow$7.14\% & $\uparrow$2.12\% & $\downarrow$0.73\% & $\downarrow$0.63\% & $\uparrow$4.84\% & $\uparrow$2.81\% \\
\bottomrule[1.2pt]
\end{tabular*}
\endgroup
\caption{Main results on five mathematical reasoning benchmarks. Scores are answer accuracy (\%). In each column within a backbone, the best completed result is in \textbf{bold}, and the second-best is underlined. The $\Delta$ row reports signed absolute differences, and the $\Delta\%$ row reports relative changes with arrows, both comparing Hi-TTRL with TTRL.}
\label{tab:main_results}
\end{table*}

Following the MCMC approximation of~\citet{karan2025reasoning}, we use a lightweight block-wise sampler to approximate the power-target hint distribution. Rather than applying MCMC to complete reasoning traces, we restrict it to short prefix hints, thereby avoiding repeated long-sequence inference calls. Let $B$ be the block size and $L$ be the number of hint blocks. The sampler grows a hint block by block, applying a small number of MCMC resampling steps at each block. After the hint is constructed, the remaining tokens are autoregressively sampled from $\pi_{\theta_{\mathrm{old}}}$.

Throughout this process, the sampler remains grounded in the old policy $\pi_{\theta_{\mathrm{old}}}$. The power target reshapes only the relative probabilities of prefixes supported by $\pi_{\theta_{\mathrm{old}}}$. At each MCMC resampling step, both prefix generation and likelihood evaluation are performed with $\pi_{\theta_{\mathrm{old}}}$. Once a hint is selected, $\pi_{\theta_{\mathrm{old}}}(\cdot\mid x,h)$ generates the remainder of the response. The resulting rollouts therefore rely solely on the old policy, without drawing on samples, probabilities, or trajectories from any external or stale policy.

\subsection{Consensus-Adaptive Hint Configuration}

Low consensus requires stronger pruning toward a common subtree, whereas high consensus requires several lightweight entry points into alternative subtrees. Accordingly, we design distinct hint configurations for different consensus levels to further steer solution-space search toward either stronger pruning or broader exploration.

We use $q_{\alpha}^{(m)}(\cdot\mid x)$ to denote the approximate power-target hint distribution restricted to prefixes of length $m$ tokens. When $c^{(1)}<\tau_{\mathrm{low}}$, the rollout group lacks a stable answer distribution, and the pseudo-label may be weakly supported. In this case, Hi-TTRL uses a sharpened power target with $\alpha_{\mathrm{low}}>1$ and samples a single long anchor hint:
\begin{equation}
\begin{aligned}
    \mathcal{H}_{\mathrm{low}}(x)
    &=
    \{h^{\mathrm{conv}}\},
    \qquad
    h^{\mathrm{conv}}
    &\sim
    q_{\alpha_{\mathrm{low}}}^{(4B)}(\cdot\mid x).
\end{aligned}
\end{equation}
The hint contains four blocks. A longer prefix imposes a stronger shared conditioning signal on the second-stage completions, shrinking the search space toward a high-likelihood reasoning subspace and encouraging answer convergence.

When $c^{(1)}>\tau_{\mathrm{high}}$, the rollout group is already overly concentrated around a dominant answer. Although this may indicate a reliable pseudo-label, it also reduces reward contrast and weakens the advantage signal. In this case, Hi-TTRL uses a flattened power target with $\alpha_{\mathrm{high}}<1$ and samples multiple short exploratory hints:
\begin{equation}
\begin{aligned}
    \mathcal{H}_{\mathrm{high}}(x)
    &=
    \{h^{\mathrm{div}}_j\}_{j=1}^{4},
    \qquad
    h^{\mathrm{div}}_j
    &\sim
    q_{\alpha_{\mathrm{high}}}^{(2B)}(\cdot\mid x).
\end{aligned}
\end{equation}
Each hint contains two blocks, and four such hints are generated. Compared with one long prefix, multiple short prefixes create several lightweight branching directions while leaving enough freedom for the policy to complete the reasoning in different ways. This promotes controlled divergence and helps recover advantage diversity.

The resulting hint pool is
\begin{equation}
    \mathcal{H}(x,c^{(1)})
    =
    \begin{cases}
    \mathcal{H}_{\mathrm{low}}(x), & c^{(1)}<\tau_{\mathrm{low}},\\
    \mathcal{H}_{\mathrm{high}}(x), & c^{(1)}>\tau_{\mathrm{high}}.
    \end{cases}
\end{equation}

For each second-stage rollout, Hi-TTRL assigns a hint from $\mathcal{H}(x,c^{(1)})$ and lets the old rollout policy complete the remaining reasoning conditioned on this hint. Thus, low consensus triggers a stronger shared prefix to promote convergence, whereas high consensus triggers multiple weaker prefixes to promote diversity.

\section{Experiments}

\subsection{Experimental Setup}
We evaluate Hi-TTRL on three open-weight backbones, Qwen2.5-Math-1.5B~\cite{yang2024qwen25math}, Qwen3-1.7B-Base~\cite{yang2025qwen3}, and Qwen3-4B-Base~\cite{yang2025qwen3}. Main experiments are conducted on multiple mathematical reasoning benchmarks, including AMC2023~\cite{hfmathaiamc23}, AIME2024~\cite{hfmathaiamc23}, GAOKAO2023-en~\cite{hfgokao2023mathen}, MATH-500~\cite{hendrycks2021math} and MINERVA~\cite{lewkowycz2022minerva}. Together, these benchmarks span diverse problem sources and difficulty levels, allowing us to evaluate whether consensus regulation remains effective across heterogeneous mathematical reasoning settings. We compare Hi-TTRL against two label-free baselines: standard TTRL~\cite{zuo2025ttrl}, which constructs pseudo-rewards from majority-voted answers, and Intuitor~\cite{zhao2025learning}, which uses the model's self-certainty as an internal reward signal. All the runs are optimized with GRPO under the same training configuration and are evaluated using both \texttt{mean@16} and \texttt{greedy@1}. We repeat each evaluation ten times and report the average results. Further details on the implementation and the standard deviations of each evaluation runs are provided in the Appendix.

\subsection{Main Results}
Based on the experimental results in Table~\ref{tab:main_results}, Hi-TTRL demonstrates consistent advantages across model families and evaluation metrics. It achieves the best average performance for all three backbones, improving over TTRL by 9.87/7.40 points on Qwen2.5-Math-1.5B, 3.34/2.70 points on Qwen3-1.7B-Base, and 2.54/1.49 points on Qwen3-4B-Base under \texttt{greedy@1}/\texttt{mean@16}, respectively. The gains are most pronounced on the smaller math-specialized backbone, where Hi-TTRL raises the average \texttt{greedy@1} score from 42.34 to 52.21 and the average \texttt{mean@16} score from 42.27 to 49.67. At the benchmark level, Hi-TTRL generally matches or surpasses TTRL, with only minor drops on Qwen3-4B-Base for AIME-2024 \texttt{mean@16} and GAOKAO2023-en. These results suggest that consensus-adaptive hinting strengthens label-free test-time adaptation by regulating rollout consensus before majority voting and reward assignment.

\subsection{Ablation Study}
\noindent\textbf{Branch-Level Ablation.}
Table~\ref{tab:ablation_branches} reports a branch-level ablation of the consensus-triggered hint mechanisms on AMC-2023. The \textbf{with-low} variant enables only the low-consensus convergence branch, the \textbf{with-high} variant enables only the high-consensus diversity branch. On Qwen2.5-Math-1.5B, both single-branch variants improve over TTRL, with \textbf{with-low} slightly stronger under \texttt{greedy@1} and \textbf{with-high} slightly stronger under \texttt{mean@16}. On Qwen3-4B-Base, \textbf{with-high} is the dominant single-branch variant, improving TTRL from 46.69/48.19 to 53.86/53.70 under \texttt{greedy@1}/\texttt{mean@16}, whereas \textbf{with-low} does not improve over TTRL on this backbone. Across both backbones, full Hi-TTRL achieves the best results in every column, showing that adaptive use of the low- and high-consensus branches is more effective than relying on either branch alone.

\begin{table}[t]
\centering
\begingroup
\setlength{\tabcolsep}{7.0 pt}
\renewcommand{\arraystretch}{1.12}
\newcommand{\greedyhead}{\shortstack{\texttt{greedy}}}
\newcommand{\avghead}{\shortstack{\texttt{mean}}}
\begin{tabular}{@{}lcccc@{}}
\toprule[1.2pt]
\multirow{2}{*}{\textbf{Model}}
& \multicolumn{2}{c}{\shortstack{Qwen2.5-Math-1.5B}}
& \multicolumn{2}{c}{\shortstack{Qwen3-4B-Base}} \\
\cmidrule(lr){2-3}\cmidrule(lr){4-5}
& \greedyhead & \avghead
& \greedyhead & \avghead \\
\midrule[1.1pt]
Baseline & 33.73 & 27.26 & 24.10 & 18.75 \\
TTRL & 44.58 & 44.88 & 46.69 & 48.19 \\
with-low & \underline{48.19} & 45.02 & 46.02 & 47.62 \\
with-high & 48.07 & \underline{45.11} & \underline{53.86} & \underline{53.70} \\
\textbf{Hi-TTRL} & \textbf{51.81} & \textbf{48.12} & \textbf{54.22} & \textbf{54.49} \\
\bottomrule[1.2pt]
\end{tabular}
\endgroup
\caption{Ablation of consensus-triggered hint branches on AMC-2023. Scores are answer accuracy (\%). The best result is in \textbf{bold}, and the second-best result is underlined.}
\label{tab:ablation_branches}
\end{table}

\noindent\textbf{Comparison with Direct Sampling Controls.}
To determine whether simpler consensus-dependent decoding adjustments can reproduce the gains of Hi-TTRL, we evaluate temperature- and top-$p$-based variants of TTRL with Qwen3-1.7B-Base. Standard TTRL samples training rollouts with temperature $T=1.0$ and top-$p=0.95$. \textbf{TTRL + Temp} keeps top-$p=0.95$ but uses $T=0.5$ for the low-consensus branch and $T=2.0$ for the high-consensus branch. \textbf{TTRL + Top-$p$} keeps $T=1.0$ but uses top-$p=0.8$ and $1.0$ for the low- and high-consensus branches, respectively. As shown in Table~\ref{tab:sampling_controls}, both direct controls improve over TTRL on AMC-2023, but remain below Hi-TTRL. On MINERVA, both direct controls degrade substantially relative to TTRL, whereas Hi-TTRL improves over TTRL on both metrics. These results indicate that branch-dependent entropy adjustment alone is insufficient; the structured prefix-level steering of Hi-TTRL provides more effective and robust consensus regulation.

\begin{table}[t]
\centering
\begingroup
\setlength{\tabcolsep}{5.0pt}
\renewcommand{\arraystretch}{1.12}
\newcommand{\greedyhead}{\shortstack{\texttt{greedy}}}
\newcommand{\avghead}{\shortstack{\texttt{mean}}}
\begin{tabular}{@{}lcccc@{}}
\toprule[1.2pt]
\multirow{2}{*}{\textbf{Method}}
& \multicolumn{2}{c}{\shortstack{AMC-2023}}
& \multicolumn{2}{c}{\shortstack{MINERVA}} \\
\cmidrule(lr){2-3}\cmidrule(lr){4-5}
& \greedyhead & \avghead
& \greedyhead & \avghead \\
\midrule[1.1pt]
Baseline & 26.51 & 22.29 & 5.88 & 8.07 \\
TTRL & 32.53 & 36.37 & \underline{30.51} & \underline{30.63} \\
TTRL + Temp. & \underline{37.35} & 36.90 & 22.98 & 23.66 \\
TTRL + Top-$p$ & \underline{37.35} & \underline{37.95} & 24.56 & 24.06 \\
\textbf{Hi-TTRL} & \textbf{38.55} & \textbf{39.76} & \textbf{34.19} & \textbf{32.74} \\
\bottomrule[1.2pt]
\end{tabular}
\endgroup
\caption{Comparison with direct sampling-control variants on Qwen3-1.7B-Base. Scores are answer accuracy (\%). The best result is in \textbf{bold}, and the second-best result is underlined.}
\label{tab:sampling_controls}
\end{table}



\begin{figure*}[t]
    \centering
    \includegraphics[width=0.95\textwidth]{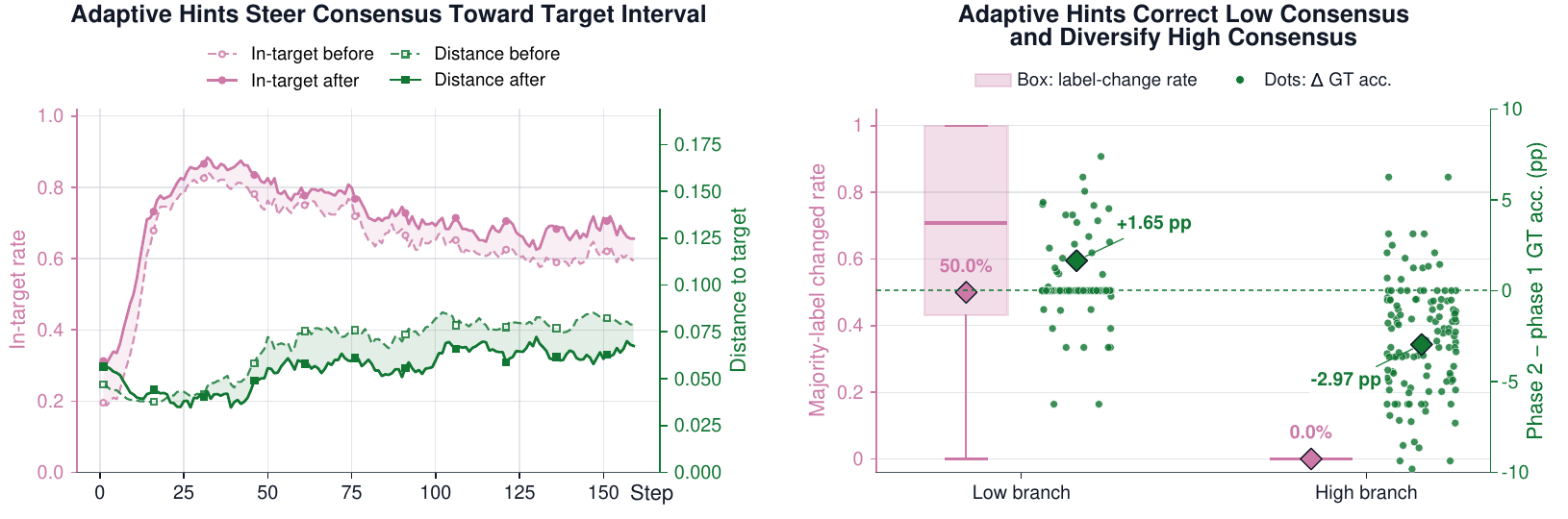}
    \caption{Mechanism analysis of adaptive hints for Qwen3-1.7B-Base on MINERVA.}
    \label{fig:adaptive_hint_mechanism}
\end{figure*}

\subsection{Mechanism Analysis}
\noindent\textbf{Consensus-Steering Effect.}
To investigate how Hi-TTRL effectively regulates consensus strength and yields more informative policy updates, we conduct a mechanism analysis on Qwen3-1.7B-Base trained on MINERVA. Specifically, at each step, we measure the \textit{in-target rate} and \textit{distance to target} over all samples before and after the hint stage. The \textit{in-target rate} measures the fraction of samples whose consensus strength lies within \([\tau_{\mathrm{low}}, \tau_{\mathrm{high}}]\), while the \textit{distance to target} is the average distance from each sample's consensus strength to this interval, with a distance of zero assigned to samples inside it. As Figure~\ref{fig:adaptive_hint_mechanism} illustrates, the in-target rate is markedly higher after the hint stage, indicating that hint-guided sampling moves a larger fraction of samples into the target interval. Meanwhile, the distance to target is markedly lower after the hint stage, showing that sample consensus strength is closer to the target interval overall. These results indicate that the hint sampler does not merely add extra rollouts, but actively steers both under- and over-concentrated groups back toward the healthy consensus range.

\noindent\textbf{Quality of Consensus-Steered Samples.}
We further examine whether consensus steering improves the quality of update samples. We assess update-sample quality using the change in majority-label rate and ground-truth accuracy from before to after hinting. As shown in Figure~\ref{fig:adaptive_hint_mechanism}, the majority label changes for a substantial fraction of low-consensus prompts, accompanied by an increase in average ground-truth rollout accuracy. In contrast, the majority label remains stable in the high-consensus branch even as rollout accuracy decreases, indicating that flattened hints introduce informative minority samples without corrupting the pseudo-label. These complementary effects show that Hi-TTRL enhances pseudo-label reliability when consensus is weak and restores advantage contrast when consensus is saturated.

\begin{figure}[t]
    \centering
    \includegraphics[width=\columnwidth]{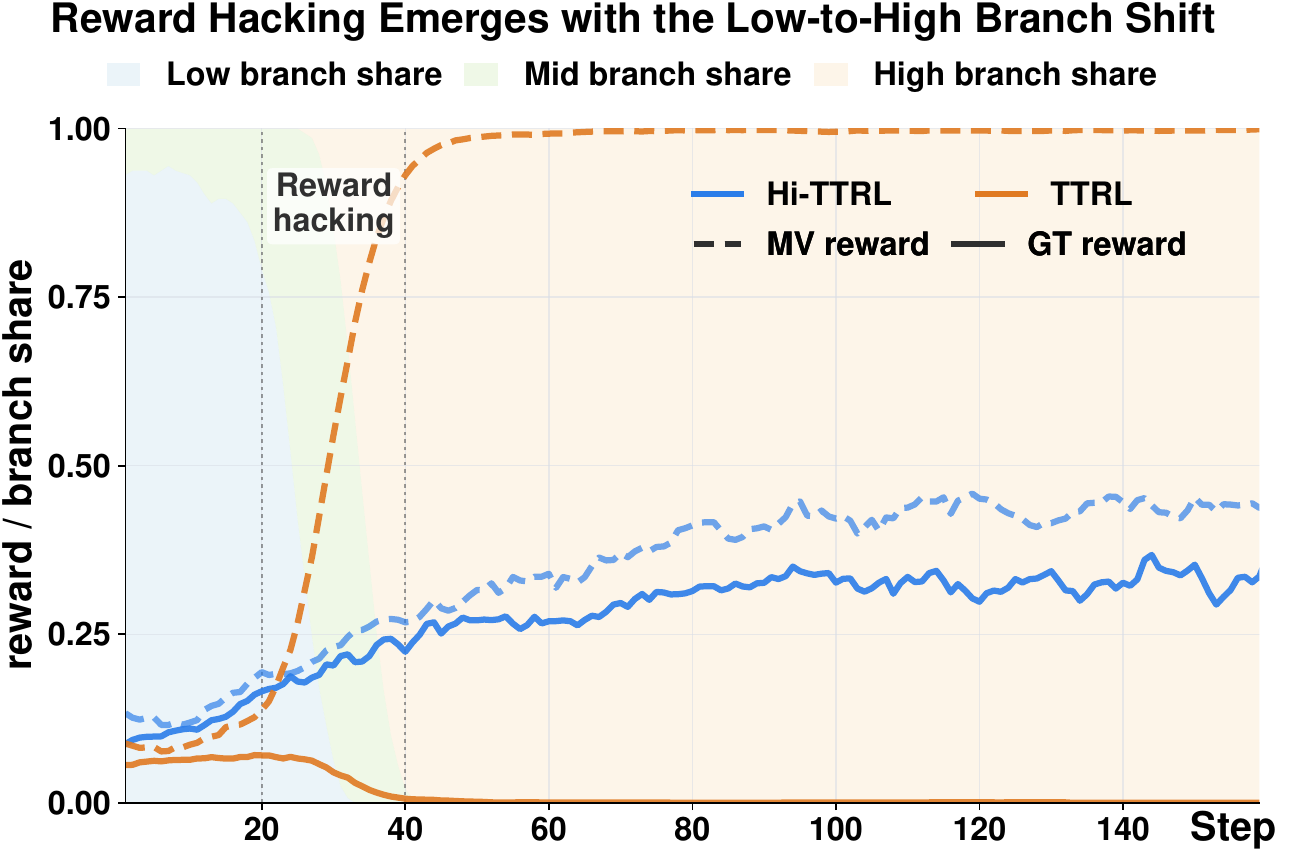}
    \caption{Reward-hacking dynamics of Qwen2.5-Math-1.5B on MINERVA. Shading shows branch shares; curves show majority-voting and ground-truth rewards.}
    \label{fig:minerva_reward_hacking}
\end{figure}

\subsection{Case Study}
To further understand how our method works, we conduct a case study. We study reward-hacking dynamics on MINERVA with Qwen2.5-Math-1.5B by tracking majority-voting (MV) reward, ground-truth (GT) reward, and the shares of low-, mid-, and high-consensus updates. As shown in Figure~\ref{fig:minerva_reward_hacking}, standard TTRL progressively migrates from low to high consensus: its MV reward saturates near one while its GT reward collapses to nearly zero. This divergence shows that weakly supported pseudo-labels can be consolidated into confident but incorrect behavior.

For Hi-TTRL, sharpened hints steer low-consensus rollouts toward a more coherent reasoning region before voting, preventing noisy majorities from becoming recurring templates. Consequently, its MV reward remains below saturation while its GT reward continues to improve. The comparison supports our central mechanism: sampling-stage consensus regulation interrupts erroneous low-to-high-consensus consolidation while preserving an informative GRPO signal.

\section{Conclusion}

We identify consensus instability as a key bottleneck in test-time reinforcement learning: weak consensus amplifies noisy pseudo-labels, while excessive consensus diminishes the advantage contrast. Hi-TTRL addresses this issue by estimating first-stage consensus and using power-target hints to induce convergence for low-consensus groups or controlled diversity for high-consensus groups. Across backbones and datasets, Hi-TTRL consistently improves over standard TTRL, with ablations, the case study, and mechanism analysis validating the effectiveness of its adaptive hint design.


\bibliography{aaai2027}

\newpage

\appendix

\section{A : Detailed Experimental Setup}
\label{app:experimental_setup}

This section provides the detailed experimental settings for the main results reported in the paper. We describe the backbones, benchmark suites, baselines, evaluation metrics, and training hyperparameters used in the main experiments.

\noindent\textbf{Models.}
We evaluate Hi-TTRL on three open-weight backbones: Qwen2.5-Math-1.5B~\cite{yang2024qwen25math}, Qwen3-1.7B-Base~\cite{yang2025qwen3}, and Qwen3-4B-Base~\cite{yang2025qwen3}. Qwen2.5-Math-1.5B represents a compact math-specialized model, while Qwen3-1.7B-Base and Qwen3-4B-Base provide lightweight and stronger general base-model settings before instruction tuning. This choice allows us to test whether consensus-strength regulation remains effective across both math-specialized and general base policies.

\noindent\textbf{Benchmarks.}
We conduct the experiments on five mathematical reasoning benchmarks: AMC2023~\cite{hfmathaiamc23}, AIME2024~\cite{hfmathaiamc23}, GAOKAO2023-en~\cite{hfgokao2023mathen}, MATH-500~\cite{hendrycks2021math}, and MINERVA~\cite{lewkowycz2022minerva}. Each benchmark is treated as an unlabeled test stream during policy optimization: the model receives only the problem statements, derives training signals from its own rollouts, and never uses ground-truth for reward construction.

\noindent\textbf{Baselines.}
Following recent TTRL evaluations~\cite{zuo2025ttrl,wang2026scope,yan2026consensus}, we compare Hi-TTRL with three baselines for each benchmark--backbone pair. The unadapted backbone is evaluated without parameter updates and therefore preserves the original policy, providing a fixed reference for quantifying the benefit introduced by adaptation. TTRL~\cite{zuo2025ttrl} samples multiple answers, uses the most frequent answer as a pseudo-label, and assigns binary rewards according to answer agreement; these self-supervised rewards are then optimized with GRPO. Intuitor~\cite{zhao2025learning} replaces answer voting with a sequence-level \emph{self-certainty} reward derived from token-averaged $\mathrm{KL}\!\left(U\,\|\,p_{\theta}\right)$; after group-wise normalization, these intrinsic rewards define the GRPO advantages.

\noindent\textbf{Evaluation Metrics.}
We report two answer-accuracy metrics: \texttt{mean@16} and \texttt{greedy@1}. For \texttt{mean@16}, each question is decoded 16 times with stochastic sampling; each generated response is scored against the benchmark answer after final-answer extraction, and the metric is the average sample accuracy over the 16 responses, further averaged over all questions. For \texttt{greedy@1}, each question is decoded once with deterministic greedy decoding and scored in the same way. We report both \texttt{mean@16} and \texttt{greedy@1} in the final evaluation. For every method--backbone--benchmark configuration and each decoding protocol, we repeat the complete evaluation ten times. Each repetition first produces one benchmark-level accuracy by aggregating over all questions and, for \texttt{mean@16}, over the 16 sampled responses per question. We report the arithmetic mean of the ten repetition-level accuracies and compute the standard deviation over the same ten values. Table~\ref{tab:full_results_std} provides the resulting mean and standard deviation for every configuration.

\noindent\textbf{Hyperparameter Configuration.}
All methods are optimized with GRPO using the same training configuration. We set the maximum prompt length to 1,024 tokens and the maximum response length to 3,072 tokens. For each prompt, the rollout pool contains 32 responses for pseudo-label voting, from which 16 responses are used for policy updates. The actor learning rate is $5\times10^{-7}$ with a cosine schedule and a warmup ratio of 0.03; the critic learning rate is $9\times10^{-6}$. During training, rollouts are sampled with temperature 1.0 and top-$p=0.95$. During evaluation, sampled decoding for \texttt{mean@16} uses temperature 0.6 and top-$p=0.95$, while \texttt{greedy@1} uses temperature 0 and top-$p=1.0$. We cap evaluation generations at 3,072 tokens with a 4,096-token context window. 

For Hi-TTRL, the consensus thresholds are set to $\tau_{\mathrm{low}}=0.25$ and $\tau_{\mathrm{high}}=0.75$. When adaptive hints are enabled, the high-consensus branch uses $\alpha_{\mathrm{high}}=0.25$, while the low-consensus branch uses $\alpha_{\mathrm{low}}=4.0$. Each hint block contains $B=128$ tokens, and we perform $N_{\mathrm{MCMC}}=10$ Metropolis--Hastings transitions after adding each block. The low-consensus branch uses four hint blocks, and the high-consensus branch uses two hint blocks with four candidate hints. All experiments are run on a single SLURM node with 4 or 8 NVIDIA L20 GPUs, depending on the backbone and benchmark.

\section{B : Full Results with Standard Deviations}
\label{app:full_results_std}

Table~\ref{tab:full_results_std} complements Table~\ref{tab:main_results} by reporting the run-to-run variability over ten repeated evaluations. Each entry gives the mean answer accuracy together with its standard deviation across the ten repetition-level scores. The upper and lower panels contain the deterministic \texttt{greedy@1} and stochastic \texttt{mean@16} evaluations, respectively. For each backbone, the blue $\Delta/\Delta\%$ row compares Hi-TTRL with TTRL: the value before the slash is the signed absolute accuracy difference, and the value after the slash is the corresponding relative change. Bold and underlined entries denote the best and second-best mean within each backbone, benchmark, and decoding protocol.

\begin{table*}
\centering
\begingroup
\setlength{\tabcolsep}{3.1pt}
\renewcommand{\arraystretch}{1.12}
\newcommand{\stddev}[1]{{\scriptstyle\,\pm\,#1}}
\newcommand{\deltapair}[1]{{\footnotesize $#1$}}
\resizebox{\textwidth}{!}{%
\begin{tabular}{>{\raggedright\arraybackslash}p{3.0cm}*{5}{>{\centering\arraybackslash}p{3.0cm}}}
\toprule[1.2pt]
\multicolumn{6}{c}{\textbf{Greedy Decoding}\enspace(\texttt{greedy@1})} \\
\cmidrule(lr){1-6}
\textbf{Models / Methods}
& \textbf{AMC-2023}
& \textbf{MATH-500}
& \textbf{AIME-2024}
& \textbf{MINERVA}
& \textbf{GAOKAO2023-en} \\
\midrule[1.1pt]
\rowcolor{MethodGray}
\multicolumn{6}{c}{\textsc{I. Math-Specialized Backbone}} \\
\rowcolor{PendingGray}
\multicolumn{6}{c}{\textsc{Qwen2.5-Math-1.5B}} \\
Baseline & $33.73\stddev{1.70}$ & $32.04\stddev{0.78}$ & \underline{$16.33\stddev{3.99}$} & $8.82\stddev{0.60}$ & $20.00\stddev{0.81}$ \\
TTRL & \underline{$44.58\stddev{0.99}$} & \underline{$73.92\stddev{0.65}$} & $13.67\stddev{1.89}$ & $16.91\stddev{0.61}$ & $62.60\stddev{0.26}$ \\
Intuitor & $37.35\stddev{1.53}$ & $70.80\stddev{0.56}$ & $11.67\stddev{1.89}$ & \underline{$25.37\stddev{0.87}$} & \underline{$64.16\stddev{0.85}$} \\
\textbf{Hi-TTRL} & $\mathbf{51.81\stddev{1.27}}$ & $\mathbf{78.00\stddev{0.56}}$ & $\mathbf{24.33\stddev{3.22}}$ & $\mathbf{36.76\stddev{0.25}}$ & $\mathbf{70.13\stddev{0.25}}$ \\
\rowcolor{TableBlue!15}
\deltapair{\Delta/\Delta\%} & \deltapair{+7.23\,/\,\uparrow16.22\%} & \deltapair{+4.08\,/\,\uparrow5.52\%} & \deltapair{+10.66\,/\,\uparrow77.98\%} & \deltapair{+19.85\,/\,\uparrow117.39\%} & \deltapair{+7.53\,/\,\uparrow12.03\%} \\
\midrule[1.1pt]
\rowcolor{MethodGray}
\multicolumn{6}{c}{\textsc{II. General Base Backbones}} \\
\rowcolor{PendingGray}
\multicolumn{6}{c}{\textsc{Qwen3-1.7B-Base}} \\
Baseline & $26.51\stddev{2.14}$ & $44.52\stddev{0.78}$ & $4.33\stddev{4.17}$ & $5.88\stddev{0.18}$ & $25.45\stddev{0.83}$ \\
TTRL & \underline{$32.53\stddev{1.52}$} & \underline{$69.80\stddev{0.42}$} & \underline{$9.67\stddev{1.05}$} & \underline{$30.51\stddev{0.68}$} & \underline{$59.40\stddev{0.32}$} \\
Intuitor & $30.12\stddev{2.11}$ & $56.80\stddev{0.75}$ & $4.33\stddev{2.83}$ & $24.26\stddev{0.46}$ & $49.87\stddev{1.03}$ \\
\textbf{Hi-TTRL} & $\mathbf{38.55\stddev{1.63}}$ & $\mathbf{71.40\stddev{0.40}}$ & $\mathbf{12.67\stddev{1.72}}$ & $\mathbf{34.19\stddev{0.78}}$ & $\mathbf{61.82\stddev{0.66}}$ \\
\rowcolor{TableBlue!15}
\deltapair{\Delta/\Delta\%} & \deltapair{+6.02\,/\,\uparrow18.51\%} & \deltapair{+1.60\,/\,\uparrow2.29\%} & \deltapair{+3.00\,/\,\uparrow31.02\%} & \deltapair{+3.68\,/\,\uparrow12.06\%} & \deltapair{+2.42\,/\,\uparrow4.07\%} \\
\midrule
\rowcolor{PendingGray}
\multicolumn{6}{c}{\textsc{Qwen3-4B-Base}} \\
Baseline & $24.10\stddev{2.26}$ & $34.72\stddev{1.22}$ & $13.33\stddev{1.05}$ & $12.13\stddev{0.63}$ & $14.55\stddev{0.49}$ \\
TTRL & $46.69\stddev{0.84}$ & \underline{$81.68\stddev{0.29}$} & \underline{$21.67\stddev{1.05}$} & \underline{$41.18\stddev{0.31}$} & $\mathbf{71.17\stddev{0.23}}$ \\
Intuitor & \underline{$51.81\stddev{0.68}$} & $74.40\stddev{0.75}$ & $9.33\stddev{3.31}$ & $29.04\stddev{0.80}$ & $69.09\stddev{0.86}$ \\
\textbf{Hi-TTRL} & $\mathbf{54.22\stddev{1.02}}$ & $\mathbf{82.76\stddev{0.16}}$ & $\mathbf{23.33\stddev{0.00}}$ & $\mathbf{44.12\stddev{0.51}}$ & \underline{$70.65\stddev{0.28}$} \\
\rowcolor{TableBlue!15}
\deltapair{\Delta/\Delta\%} & \deltapair{+7.53\,/\,\uparrow16.13\%} & \deltapair{+1.08\,/\,\uparrow1.32\%} & \deltapair{+1.66\,/\,\uparrow7.66\%} & \deltapair{+2.94\,/\,\uparrow7.14\%} & \deltapair{-0.52\,/\,\downarrow0.73\%} \\
\midrule[1.2pt]
\multicolumn{6}{c}{\textbf{Sampled Decoding}\enspace(\texttt{mean@16})} \\
\cmidrule(lr){1-6}
\textbf{Models / Methods}
& \textbf{AMC-2023}
& \textbf{MATH-500}
& \textbf{AIME-2024}
& \textbf{MINERVA}
& \textbf{GAOKAO2023-en} \\
\midrule[1.1pt]
\rowcolor{MethodGray}
\multicolumn{6}{c}{\textsc{I. Math-Specialized Backbone}} \\
\rowcolor{PendingGray}
\multicolumn{6}{c}{\textsc{Qwen2.5-Math-1.5B}} \\
Baseline & $27.26\stddev{1.06}$ & $34.32\stddev{0.31}$ & $9.58\stddev{0.62}$ & $8.46\stddev{0.30}$ & $21.79\stddev{0.37}$ \\
TTRL & \underline{$44.88\stddev{0.69}$} & \underline{$71.35\stddev{0.32}$} & \underline{$15.88\stddev{0.88}$} & $17.14\stddev{0.32}$ & \underline{$62.11\stddev{0.43}$} \\
Intuitor & $36.45\stddev{0.83}$ & $67.63\stddev{0.30}$ & $9.29\stddev{0.88}$ & \underline{$22.04\stddev{0.11}$} & $58.86\stddev{0.22}$ \\
\textbf{Hi-TTRL} & $\mathbf{48.12\stddev{0.69}}$ & $\mathbf{76.86\stddev{0.36}}$ & $\mathbf{18.25\stddev{0.79}}$ & $\mathbf{35.96\stddev{0.36}}$ & $\mathbf{69.17\stddev{0.24}}$ \\
\rowcolor{TableBlue!15}
\deltapair{\Delta/\Delta\%} & \deltapair{+3.24\,/\,\uparrow7.22\%} & \deltapair{+5.51\,/\,\uparrow7.72\%} & \deltapair{+2.37\,/\,\uparrow14.92\%} & \deltapair{+18.82\,/\,\uparrow109.80\%} & \deltapair{+7.06\,/\,\uparrow11.37\%} \\
\midrule[1.1pt]
\rowcolor{MethodGray}
\multicolumn{6}{c}{\textsc{II. General Base Backbones}} \\
\rowcolor{PendingGray}
\multicolumn{6}{c}{\textsc{Qwen3-1.7B-Base}} \\
Baseline & $22.29\stddev{0.84}$ & $39.07\stddev{0.42}$ & $2.08\stddev{0.64}$ & $8.07\stddev{0.12}$ & $23.28\stddev{0.41}$ \\
TTRL & \underline{$36.37\stddev{0.55}$} & \underline{$69.17\stddev{0.19}$} & \underline{$7.50\stddev{0.97}$} & \underline{$30.63\stddev{0.36}$} & \underline{$59.38\stddev{0.22}$} \\
Intuitor & $26.51\stddev{1.18}$ & $56.65\stddev{0.26}$ & $4.58\stddev{1.02}$ & $22.52\stddev{0.14}$ & $48.17\stddev{0.33}$ \\
\textbf{Hi-TTRL} & $\mathbf{39.76\stddev{0.43}}$ & $\mathbf{71.26\stddev{0.12}}$ & $\mathbf{11.25\stddev{0.44}}$ & $\mathbf{32.74\stddev{0.19}}$ & $\mathbf{61.56\stddev{0.26}}$ \\
\rowcolor{TableBlue!15}
\deltapair{\Delta/\Delta\%} & \deltapair{+3.39\,/\,\uparrow9.32\%} & \deltapair{+2.09\,/\,\uparrow3.02\%} & \deltapair{+3.75\,/\,\uparrow50.00\%} & \deltapair{+2.11\,/\,\uparrow6.89\%} & \deltapair{+2.18\,/\,\uparrow3.67\%} \\
\midrule
\rowcolor{PendingGray}
\multicolumn{6}{c}{\textsc{Qwen3-4B-Base}} \\
Baseline & $18.75\stddev{0.51}$ & $31.89\stddev{0.30}$ & $7.50\stddev{0.99}$ & $13.40\stddev{0.12}$ & $12.73\stddev{0.33}$ \\
TTRL & $48.19\stddev{0.41}$ & \underline{$81.67\stddev{0.12}$} & $\mathbf{21.04\stddev{0.43}}$ & \underline{$42.37\stddev{0.26}$} & $\mathbf{71.10\stddev{0.06}}$ \\
Intuitor & \underline{$53.09\stddev{0.46}$} & $74.92\stddev{0.23}$ & $9.52\stddev{0.95}$ & $29.50\stddev{0.08}$ & $68.39\stddev{0.35}$ \\
\textbf{Hi-TTRL} & $\mathbf{54.49\stddev{0.46}}$ & $\mathbf{82.57\stddev{0.05}}$ & \underline{$20.83\stddev{0.34}$} & $\mathbf{43.27\stddev{0.25}}$ & \underline{$70.65\stddev{0.17}$} \\
\rowcolor{TableBlue!15}
\deltapair{\Delta/\Delta\%} & \deltapair{+6.30\,/\,\uparrow13.07\%} & \deltapair{+0.90\,/\,\uparrow1.10\%} & \deltapair{-0.21\,/\,\downarrow1.00\%} & \deltapair{+0.90\,/\,\uparrow2.12\%} & \deltapair{-0.45\,/\,\downarrow0.63\%} \\
\bottomrule[1.2pt]
\end{tabular}%
}
\endgroup
\caption{Full \texttt{greedy@1} and \texttt{mean@16} results with standard deviations on five mathematical reasoning benchmarks. The upper and lower panels report the two decoding protocols, respectively. Entries are mean answer accuracy (\%) $\pm$ standard deviation over ten repeated evaluations. In each column within a backbone and decoding protocol, the best mean is in \textbf{bold}, and the second-best mean is underlined. In each blue $\Delta/\Delta\%$ row, the values before and after the slash report the signed absolute and relative changes of Hi-TTRL over TTRL, respectively.}
\label{tab:full_results_std}
\end{table*}

\section{C : Algorithmic Details}
\label{app:mcmc_details}

\subsection{Relation to Power Sampling}

Our MCMC implementation directly follows the autoregressive power-sampling approximation of~\citet{karan2025reasoning}. Specifically, we retain its sequence-level power target, progressive block-wise intermediate targets, and random-suffix Metropolis--Hastings (MH) transitions. The distinction is how this sampler is used. The original method applies power sampling to a complete inference-time response and studies sharpening with $\alpha>1$. Hi-TTRL restricts the MCMC approximation to a short prefix $h$, allows either $\alpha_{\mathrm{low}}>1$ or $\alpha_{\mathrm{high}}<1$, and invokes the sampler only when $c^{(1)}$ lies outside $[\tau_{\mathrm{low}},\tau_{\mathrm{high}}]$. The sampled prefix is then used as a hint, and the remainder of the rollout is generated normally by $\pi_{\theta_{\mathrm{old}}}$. Thus, the MCMC transition rule is inherited from power sampling, whereas consensus-adaptive routing, flattening, prefix truncation, and hint-conditioned policy updates are specific to Hi-TTRL.

\subsection{Block-wise Metropolis--Hastings Transitions}

Given a query $x$ and a branch-specific exponent $\alpha$, the sampler produces a hint $h$ of at most $M=LB$ tokens. It initializes $h$ as an empty sequence and constructs the hint in $L$ stages. At stage $k$, the current hint contains $(k-1)B$ tokens. The sampler first appends $B$ tokens drawn from $\pi_{\theta_{\mathrm{old}}}$, obtaining an initial state of length $kB$. It then applies $N_{\mathrm{MCMC}}$ updates to this state. In each update, the sampler chooses a position $m$, keeps the tokens before $m$, and autoregressively regenerates every token from $m$ through $kB$. The resulting sequence replaces the current state only if it passes the MH acceptance test for the power target $\pi_{\theta_{\mathrm{old}}}(h\mid x)^{\alpha}$; otherwise, the current state is retained. After these updates, the resulting $kB$-token state becomes the starting prefix for stage $k+1$. The state obtained after stage $L$ is returned as the final hint.

This construction adapts the progressive power-sampling procedure of~\citet{karan2025reasoning}. Instead of initializing MCMC with an independently sampled full-length hint, each stage reuses the prefix refined at the preceding stage and samples only one new block before applying MH updates. Both Hi-TTRL routes use this same transition mechanism: $\alpha$ determines whether the target distribution is sharpened or flattened, while the branch-specific hint lengths and reuse strategies are specified below.

\noindent\textbf{Intermediate targets and warm starts.}
For a prefix containing $k$ blocks, define its old-policy log-likelihood as
\begin{equation}
    \ell_k(h\mid x)
    =
    \sum_{t=1}^{kB}
    \log \pi_{\theta_{\mathrm{old}}}(h_t\mid x,h_{<t}).
\end{equation}
Consistent with Eq.~\eqref{eq:power_hint_main}, the $k$-th unnormalized intermediate target is
\begin{equation}
    q_{\alpha,k}(h_{1:kB}\mid x)
    \propto
    \exp\!\left\{\alpha\ell_k(h\mid x)\right\}
    =
    \pi_{\theta_{\mathrm{old}}}(h_{1:kB}\mid x)^{\alpha}.
    \label{eq:intermediate_power_target}
\end{equation}
These targets form a length-increasing path from the empty sequence to the desired $M$-token hint distribution. To initialize the chain for $q_{\alpha,k}$, we retain the accepted first $(k-1)B$ tokens and draw only the next $B$ tokens autoregressively from $\pi_{\theta_{\mathrm{old}}}$. Thus, the previous block supplies the prefix of the new state, while ordinary old-policy sampling supplies its newly added suffix.

\noindent\textbf{Suffix proposals and MH correction.}
Given the initialized $kB$-token state $h$, each MH transition samples a resampling position $m$ uniformly from $\{1,\ldots,kB\}$. It keeps $h_{<m}$ fixed and proposes a new suffix from the old policy,
\begin{equation}
    h'_{m:kB}
    \sim
    \pi_{\theta_{\mathrm{old}}}
    (\cdot\mid x,h_{<m}),
    \qquad h'_{<m}=h_{<m}.
    \label{eq:mcmc_suffix_proposal}
\end{equation}
Conditioned on the sampled position $m$, the proposal kernel has support only on sequences satisfying $h'_{<m}=h_{<m}$. On this support, the forward and reverse suffix-proposal probabilities are
\begin{equation}
\begin{aligned}
    r_m(h'\mid h,x)
    &=
    \prod_{t=m}^{kB}
    \pi_{\theta_{\mathrm{old}}}
    (h'_t\mid x,h'_{<t}),\\
    r_m(h\mid h',x)
    &=
    \prod_{t=m}^{kB}
    \pi_{\theta_{\mathrm{old}}}
    (h_t\mid x,h_{<t}).
\end{aligned}
\label{eq:mcmc_suffix_kernel}
\end{equation}
Because $r_m$ is conditioned on $m$, these expressions do not include the factor $1/(kB)$ for selecting the resampling position; that factor is identical in the forward and reverse transitions and would cancel in the MH ratio. The candidate is accepted with probability
\begin{equation}
\begin{aligned}
    A_{\alpha}(h',h)
    =\min\!\left\{1,
    \frac{q_{\alpha,k}(h'\mid x)\,
          r_m(h\mid h',x)}
         {q_{\alpha,k}(h\mid x)\,
          r_m(h'\mid h,x)}
    \right\}.
    \label{eq:mcmc_acceptance_general}
\end{aligned}
\end{equation}
Here, we use $u\sim\mathrm{Uniform}(0,1)$ as an independent auxiliary draw used for the accept--reject decision: accept $h'$ if $u\leq A_{\alpha}(h',h)$, and otherwise retain $h$. Since $h$ and $h'$ share the same prefix before $m$, their full old-policy likelihood ratio reduces to the likelihood ratio of the two suffixes. Moreover, the forward and reverse suffix proposals use that same old policy. The proposal correction therefore cancels one power of the old-policy likelihood, reducing Eq.~\eqref{eq:mcmc_acceptance_general} to
\begin{equation}
    \log A_{\alpha}(h',h)
    =
    \min\!\left\{0,
    (\alpha-1)
    \left[\ell_k(h'\mid x)-\ell_k(h\mid x)\right]
    \right\}.
    \label{eq:mcmc_acceptance_simplified}
\end{equation}

\noindent\textbf{Routing behavior and implementation.}
We evaluate Eq.~\eqref{eq:mcmc_acceptance_simplified} in log space for numerical stability. Its behavior is directly controlled by $\alpha$: when $\alpha>1$, higher-likelihood proposals are favored, concentrating hints in high-probability regions of the old policy; when $\alpha<1$, likelihood-decreasing moves are favored, flattening the hint distribution and encouraging exploration. At $\alpha=1$, every proposal is accepted, recovering ordinary old-policy sampling.

For each block, we apply $N_{\mathrm{MCMC}}$ transitions and carry the resulting state---including an unchanged state after a rejection---into the next block. We do not discard a separate burn-in segment, because the progressive initialization provides the warm start at each sequence length. After block $L$, the final state is returned as one hint. Independent chains produce the four high-consensus hints, which are assigned as evenly as possible across the $G/2$ second-stage rollouts; the single low-consensus anchor is shared by all $G/2$ second-stage rollouts. In our experiments, $B=128$ and $N_{\mathrm{MCMC}}=10$, with $L=4$ for the low-consensus branch and $L=2$ for the high-consensus branch. Algorithm~\ref{alg:mcmc_hint} gives the complete block-wise procedure.

\begin{algorithm}[t]
\caption{Power-Target MCMC Hint Sampler}
\label{alg:mcmc_hint}
\begin{algorithmic}[1]
\STATE \textbf{Input:} query $x$, old policy $\pi_{\theta_{\mathrm{old}}}$, exponent $\alpha$, block size $B$, number of blocks $L$, transitions $N_{\mathrm{MCMC}}$
\STATE Initialize the accepted prefix $h\leftarrow\varnothing$
\FOR{$k=1$ to $L$}
    \STATE Extend $h$ by $B$ tokens sampled from $\pi_{\theta_{\mathrm{old}}}(\cdot\mid x,h)$
    \FOR{$s=1$ to $N_{\mathrm{MCMC}}$}
        \STATE Sample $m\sim\mathrm{Uniform}\{1,\ldots,kB\}$
        \STATE Set $h'_{<m}=h_{<m}$ and resample $h'_{m:kB}$ from $\pi_{\theta_{\mathrm{old}}}(\cdot\mid x,h_{<m})$
        \STATE Compute $\log A_{\alpha}(h',h)$ using Eq.~\eqref{eq:mcmc_acceptance_simplified}
        \STATE Draw $u\sim\mathrm{Uniform}(0,1)$
        \IF{$\log u\leq \log A_{\alpha}(h',h)$}
            \STATE Accept the proposal: $h\leftarrow h'$
        \ENDIF
    \ENDFOR
\ENDFOR
\STATE \textbf{return} hint $h$
\end{algorithmic}
\end{algorithm}

\subsection{Overall Training Procedure}

Algorithm~\ref{alg:hittrl} summarizes the complete Hi-TTRL training step. Hi-TTRL first samples a half group of rollouts and uses its consensus strength to determine whether a direct update is appropriate or a second sampling stage is required. For an out-of-interval group, the MCMC sampler targeting the selected power distribution generates either anchor or exploratory hints, which guide the remaining rollouts toward the desired consensus regime. The resulting adaptive rollout group is then used for the final majority vote, pseudo-reward computation, advantage normalization, and clipped GRPO update.

\begin{algorithm}[H]
\caption{Hi-TTRL Training Step}
\label{alg:hittrl}
\begin{algorithmic}[1]
\STATE \textbf{Input:} query $x$, old policy $\pi_{\theta_{\mathrm{old}}}$, rollout budget $G$, thresholds $\tau_{\mathrm{low}},\tau_{\mathrm{high}}$, exponents $\alpha_{\mathrm{low}},\alpha_{\mathrm{high}}$
\STATE Sample first-stage rollouts $\mathcal{Y}^{(1)}=\{y_i\}_{i=1}^{G/2}$ from $\pi_{\theta_{\mathrm{old}}}(\cdot\mid x)$
\STATE Extract answers, compute intermediate majority vote, and compute $c^{(1)}=c(\mathcal{Y}^{(1)})$
\IF{$c^{(1)}\in[\tau_{\mathrm{low}},\tau_{\mathrm{high}}]$}
    \STATE Set $\mathcal{Y}_{\mathrm{upd}}(x)=\mathcal{Y}^{(1)}$
\ELSE
    \IF{$c^{(1)}<\tau_{\mathrm{low}}$}
        \STATE Set $\alpha=\alpha_{\mathrm{low}}>1$ and sample anchor hints with the power-target MCMC sampler
    \ELSE
        \STATE Set $\alpha=\alpha_{\mathrm{high}}<1$ and sample exploratory hints with the power-target MCMC sampler
    \ENDIF
    \STATE Complete each hint with $\pi_{\theta_{\mathrm{old}}}(\cdot\mid x,h)$ to obtain $\mathcal{Y}^{(2)}_{\mathrm{hint}}=\{y_i\}_{i=G/2+1}^{G}$
    \STATE Set $\mathcal{Y}_{\mathrm{upd}}(x)=\mathcal{Y}^{(1)}\cup\mathcal{Y}^{(2)}_{\mathrm{hint}}$
\ENDIF
\STATE Compute the final majority vote, pseudo-rewards, and normalized GRPO advantages on $\mathcal{Y}_{\mathrm{upd}}(x)$
\STATE Recompute old-policy log probabilities for all hint and completion tokens
\STATE Update $\theta$ with the standard clipped GRPO surrogate over $\mathcal{Y}_{\mathrm{upd}}(x)$
\end{algorithmic}
\end{algorithm}

\section{D : Sample-Budget Control}
\label{app:sample_count_control}

This section provides an additional control experiment on AMC-2023 to test whether the gains of Hi-TTRL come merely from using more samples for policy-gradient updates. In Hi-TTRL, samples routed to the normal branch update the policy with 16 rollouts, while samples that trigger the adaptive-hint branch update the policy with 32 rollouts after hint-conditioned completion. This mixed update size raises a possible alternative explanation: the improvement may come from increasing the number of update rollouts rather than from adaptive hinting itself.

To isolate this factor, we compare Hi-TTRL with two standard TTRL variants that do not use adaptive hints. The first variant, TTRL$_{32/16}$, follows the default setting: it uses 32 rollouts for majority voting and 16 rollouts for GRPO updates. The second variant, TTRL$_{64/32}$, doubles both the voting and update budgets, using 64 rollouts for majority voting and 32 rollouts for GRPO updates. Thus, TTRL$_{64/32}$ gives standard TTRL at least as many update rollouts as the adaptive-hint branch of Hi-TTRL, but without changing the sampling distribution through hints. All methods use the same optimization configuration and evaluation protocol.

\begin{table}[t]
\centering
\begingroup
\setlength{\tabcolsep}{2.0pt}
\renewcommand{\arraystretch}{1.12}
\newcommand{\greedyhead}{\shortstack{\texttt{greedy}}}
\newcommand{\avghead}{\shortstack{\texttt{mean}}}
\begin{tabular}{@{}lcccc@{}}
\toprule[1.2pt]
\multirow{2}{*}{\textbf{Method}}
& \multicolumn{2}{c}{\shortstack{Qwen2.5-Math-1.5B}}
& \multicolumn{2}{c}{\shortstack{Qwen3-1.7B-Base}} \\
\cmidrule(lr){2-3}\cmidrule(lr){4-5}
& \greedyhead & \avghead
& \greedyhead & \avghead \\
\midrule[1.1pt]
Baseline & 33.73 & 27.26 & 26.51 & 22.29 \\
TTRL$_{32/16}$ & 44.58 & 44.88 & 32.53 & \underline{36.37} \\
TTRL$_{64/32}$ & \underline{49.40} & \underline{46.91} & \underline{36.27} & 36.07 \\
\textbf{Hi-TTRL} & \textbf{51.81} & \textbf{48.12} & \textbf{38.55} & \textbf{39.76} \\
\bottomrule[1.2pt]
\end{tabular}
\endgroup
\caption{Control experiment on rollout and update sample counts on AMC-2023. TTRL$_{32/16}$ uses 32 voting rollouts and 16 update rollouts, while TTRL$_{64/32}$ uses 64 voting rollouts and 32 update rollouts. Hi-TTRL uses the normal adaptive setting, where non-triggered samples use 16 update rollouts and triggered adaptive-hint samples use 32 update rollouts. Scores are answer accuracy (\%). In each column, the best result is in \textbf{bold}, and the second-best result is underlined.}
\label{tab:sample_count_control}
\end{table}

Table~\ref{tab:sample_count_control} shows that increasing the TTRL sample budget alone does not explain the gains of Hi-TTRL. On Qwen2.5-Math-1.5B, expanding standard TTRL from TTRL$_{32/16}$ to TTRL$_{64/32}$ improves AMC-2023 performance from 44.58/44.88 to 49.40/46.91 under \texttt{greedy@1}/\texttt{mean@16}, confirming that a larger rollout budget can strengthen standard TTRL. However, Hi-TTRL further improves the result to 51.81/48.12, outperforming the larger-budget TTRL$_{64/32}$ setting on both metrics, even though Hi-TTRL does not use 32 update rollouts for every problem. On Qwen3-1.7B-Base, TTRL$_{64/32}$ reaches 36.27/36.07, while Hi-TTRL further improves the result to 38.55/39.76.

These results indicate that Hi-TTRL's advantage is not simply a consequence of larger policy-gradient batches. Rather, adaptive hinting changes which reasoning prefixes are completed before voting and reward assignment, allowing the method to regulate the rollout distribution more effectively than standard TTRL with a uniformly larger sample budget.

\section{E : Experiment Scaling to Larger Models}
\label{app:qwen25math7b_dynamics}

This section examines whether Hi-TTRL generalizes beyond the model scales covered by the main experiments. Due to computational constraints, the main results focus on backbones up to the 4B scale, including Qwen2.5-Math-1.5B, Qwen3-1.7B-Base, and Qwen3-4B-Base. To further test scalability, we introduce an additional 7B-scale model, Qwen2.5-Math-7B, on AMC-2023. We also include the Qwen2.5-Math-1.5B TTRL and Hi-TTRL trajectories as a lower-scale reference under the same validation protocol. This comparison allows us to assess whether the adaptive-hint mechanism remains effective when moving from the sub-4B setting to a larger 7B-scale backbone.

\begin{figure}[t]
    \centering
    \includegraphics[width=\columnwidth]{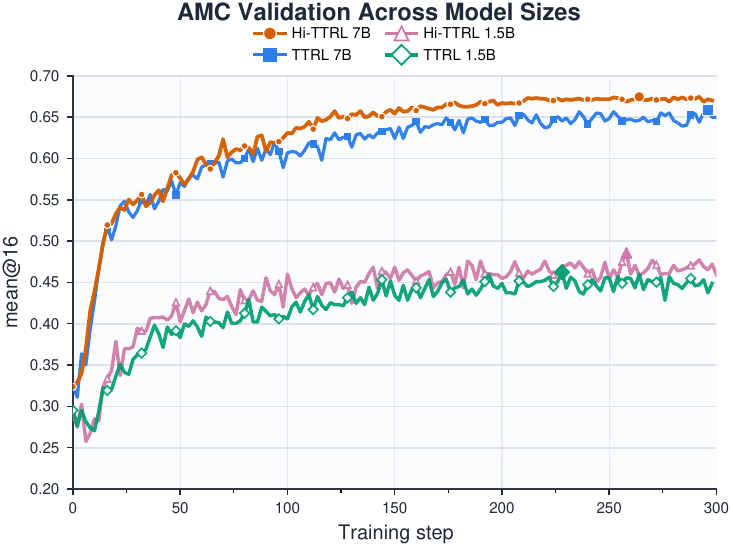}
    \caption{Generalization to a 7B-scale model on AMC-2023. Curves report validation \texttt{mean@16} during test-time adaptation, including Qwen2.5-Math-1.5B TTRL and Hi-TTRL trajectories together with the Qwen2.5-Math-7B result.}
    \label{fig:qwen25math7b_dynamics}
\end{figure}

Figure~\ref{fig:qwen25math7b_dynamics} reports the validation \texttt{mean@16} trajectories during adaptation. The key purpose of this figure is to verify that Hi-TTRL is not limited to the model scales used in the main-result table. On Qwen2.5-Math-1.5B, the comparison with TTRL gives a reference point for the benefit of replacing passive majority-voting updates with consensus-adaptive hinting. The Qwen2.5-Math-7B result then extends the evaluation to a larger model scale, showing that the same adaptation strategy can still provide useful gains beyond the model scales of up to 4B parameters studied in the main experiments.

Overall, this result strengthens the model-scale generalization claim of Hi-TTRL. The 7B experiment uses the same design principle as the smaller-scale experiments: estimate rollout consensus, intervene only when the group falls outside the target consensus interval, and use hint-conditioned completions before reward assignment. Its effectiveness on Qwen2.5-Math-7B indicates that Hi-TTRL's gains are not an artifact of a particular small backbone. Together with the sample-budget control in Appendix D, this supports the interpretation that adaptive rollout steering remains beneficial as model scale increases.

\section{F : Generalization to Non-Mathematical Scientific Reasoning}
\label{app:gpqa}

\noindent\textbf{Experimental Setup.}
To evaluate whether consensus-adaptive test-time reinforcement learning extends beyond mathematical problem solving, we conduct an additional experiment on GPQA~\cite{rein2024gpqa}, a graduate-level multiple-choice benchmark written and validated by domain experts in biology, physics, and chemistry. We use the two general-purpose backbones, Qwen3-1.7B-Base and Qwen3-4B-Base, and compare the unadapted backbone, TTRL, Intuitor, and Hi-TTRL. As in the main experiments, GPQA is treated as an unlabeled test stream during policy optimization: the methods receive the question and answer choices but not the correct choice or expert explanation, and ground-truth answers are used only for final evaluation. We retain the GRPO, rollout, adaptive-hint, and decoding configurations described in Appendix A, and report answer accuracy under \texttt{greedy@1} and \texttt{mean@16}.

\begin{table}[t]
\centering
\begingroup
\setlength{\tabcolsep}{5.0pt}
\renewcommand{\arraystretch}{1.12}
\newcommand{\greedyhead}{\shortstack{\texttt{greedy}}}
\newcommand{\avghead}{\shortstack{\texttt{mean}}}
\begin{tabular}{@{}lcccc@{}}
\toprule[1.2pt]
\multirow{2}{*}{\textbf{Method}}
& \multicolumn{2}{c}{\shortstack{Qwen3-1.7B-Base}}
& \multicolumn{2}{c}{\shortstack{Qwen3-4B-Base}} \\
\cmidrule(lr){2-3}\cmidrule(lr){4-5}
& \greedyhead & \avghead
& \greedyhead & \avghead \\
\midrule[1.1pt]
Baseline & 1.97 & 2.41 & 2.53 & 2.45 \\
TTRL & \underline{23.69} & \underline{23.63} & \underline{37.42} & \underline{38.07} \\
Intuitor & 6.87 & 7.01 & 8.89 & 9.42 \\
\textbf{Hi-TTRL} & \textbf{28.79} & \textbf{28.24} & \textbf{44.95} & \textbf{40.15} \\
\bottomrule[1.2pt]
\end{tabular}
\endgroup
\caption{Answer accuracy (\%) on GPQA under \texttt{greedy@1} and \texttt{mean@16}. The best result in each column is in \textbf{bold}, and the second-best result is underlined.}
\label{tab:gpqa_results}
\end{table}

\noindent\textbf{Results.}
Table~\ref{tab:gpqa_results} shows that Hi-TTRL achieves the highest accuracy for both backbones and decoding protocols. Relative to TTRL, it improves Qwen3-1.7B-Base by 5.10 and 4.61 percentage points under \texttt{greedy@1} and \texttt{mean@16}, respectively, and improves Qwen3-4B-Base by 7.53 and 2.08 points. The largest gain occurs with Qwen3-4B-Base under greedy decoding, where accuracy increases from 37.42\% to 44.95\%. TTRL is the strongest baseline in every setting, while Hi-TTRL consistently adds further gains, indicating that regulating rollout consensus before pseudo-label construction is also beneficial for expert-level scientific question answering. Although this experiment covers only GPQA and two general-purpose backbones, it provides initial evidence that the mechanism is not confined to mathematical benchmarks.

\begin{figure*}[t]
    \centering
    \includegraphics[width=0.9\textwidth]{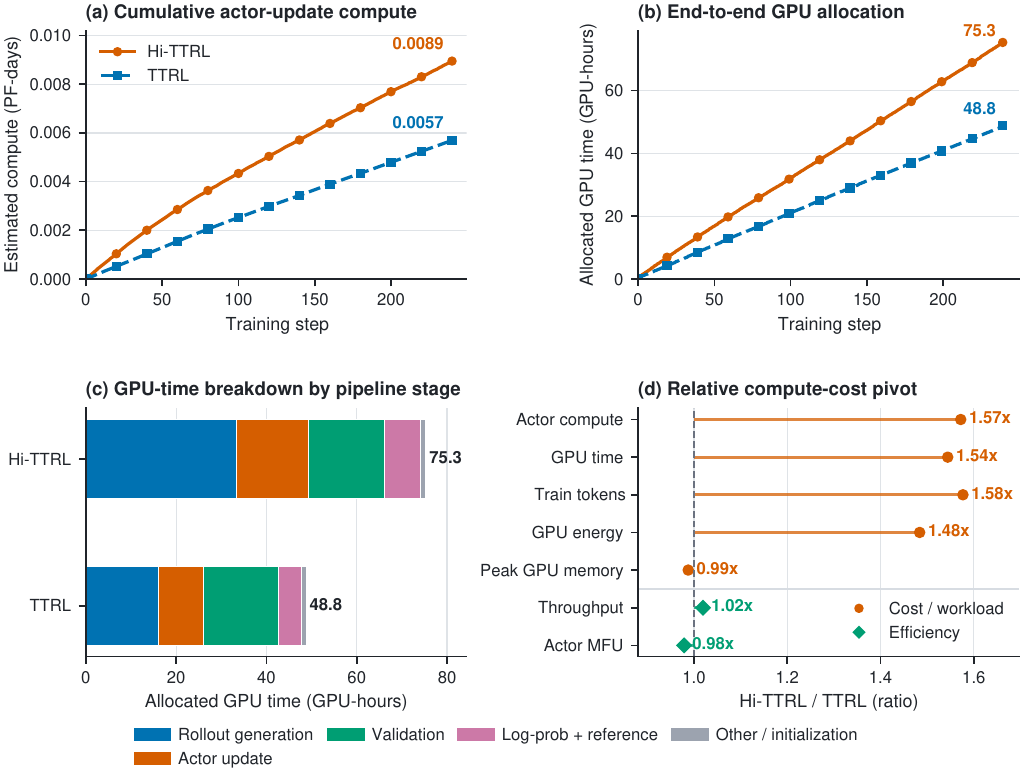}
    \caption{Computational cost comparison between Hi-TTRL and standard TTRL on AIME-2024 with Qwen2.5-Math-1.5B. The panels report cumulative actor-update compute, end-to-end GPU allocation, GPU-time breakdown by pipeline stage, and relative cost/efficiency ratios.}
    \label{fig:compute_cost}
\end{figure*}

\section{G : Computational Cost Analysis}
\label{app:compute_cost}

This section measures the computational overhead of Hi-TTRL relative to standard TTRL on AIME-2024 with Qwen2.5-Math-1.5B. Because Hi-TTRL invokes power-target hint generation and second-stage completion only when first-stage consensus falls outside the target interval, its overhead depends on the trigger rate rather than uniformly doubling the pipeline. As shown in Figure~\ref{fig:compute_cost}, Hi-TTRL uses 75.3 GPU-hours compared with 48.8 GPU-hours for TTRL, a $1.54\times$ increase in end-to-end allocated GPU time.

Panel (a) isolates actor-update compute: Hi-TTRL reaches 0.0089 PF-days versus 0.0057 PF-days for TTRL, corresponding to $1.57\times$. This ratio closely tracks the $1.58\times$ increase in training tokens, and both curves grow approximately linearly. Panel (b) includes the full training and evaluation pipeline. Its similar $1.54\times$ ratio indicates that the additional cost primarily reflects computation on hint-augmented rollouts rather than idle time or scheduling overhead.

Panels (c) and (d) further localize this overhead. The additional GPU time is concentrated in rollout generation, actor update, and log-probability/reference computation, while initialization remains small. Relative to TTRL, Hi-TTRL uses $1.57\times$ actor compute, $1.54\times$ allocated GPU time, $1.58\times$ training tokens, and $1.48\times$ GPU energy. In contrast, peak memory ($0.99\times$), throughput ($1.02\times$), and actor MFU ($0.98\times$) remain nearly unchanged. Hi-TTRL therefore increases total computation without materially reducing hardware efficiency. Together with the sample-budget control in Appendix D, this result attributes the added cost to selective consensus-adaptive sampling rather than uniform or inefficient scaling of the entire pipeline.

\section{H : Preliminary-Experiment Metrics}
\label{app:preliminary_metrics}

Figure~\ref{fig:preliminary_experiment} tracks four branch-level diagnostics during standard TTRL training. At each training step, every prompt is assigned to the low-, mid-, or high-consensus branch according to the consensus strength of its rollout group. The 16 rollouts selected from that group for the GRPO update inherit the same branch label, and the following metrics are computed separately for each branch.

\noindent\textbf{Branch Sample Count.}
This metric counts the update rollouts assigned to a branch at a given training step. Since every prompt contributes the same number of update rollouts, it also reflects how frequently prompt groups fall into each consensus range over the course of training.

\noindent\textbf{Maximum Positive Advantage.}
This metric records the largest positive normalized GRPO advantage among the update rollouts in a branch. It captures the strongest probability-increasing signal produced by that branch and therefore indicates how aggressively a single rewarded rollout can influence the policy update.

\noindent\textbf{Wrong-Promoted Rate.}
This metric measures the fraction of branch-assigned update rollouts that receive a positive normalized advantage despite having an incorrect final answer under the ground-truth annotation. It quantifies how often the majority-voting reward promotes an erroneous rollout.

\noindent\textbf{Ground-Truth Branch Accuracy.}
This metric measures the fraction of update rollouts in a branch whose extracted final answers match the ground-truth answers, regardless of the signs of their advantages. It reflects the underlying answer quality of each consensus branch. Ground-truth annotations are used only to compute the wrong-promoted rate and branch accuracy for this diagnostic analysis; they are never used to construct rewards or update the policy.

\end{document}